\documentclass{article}

\PassOptionsToPackage{numbers, compress}{natbib}

\usepackage[final, nonatbib]{neurips_2022}

\usepackage[utf8]{inputenc} 
\usepackage[T1]{fontenc}    
\usepackage{hyperref}       
\usepackage{url}            
\usepackage{booktabs}       
\usepackage{amsfonts}       
\usepackage{nicefrac}       
\usepackage{microtype}      
\usepackage{xcolor}         

\usepackage{enumitem}       
\setlist[itemize]{noitemsep, topsep=0pt}

\usepackage{multirow}
\usepackage{natbib}
\usepackage{wrapfig}

\usepackage{amsmath}
\usepackage{amssymb}
\usepackage{mathtools}
\usepackage{amsthm}
\usepackage{bm}

\theoremstyle{plain}
\newtheorem{theorem}{Theorem}[section]

\theoremstyle{definition}
\newtheorem{definition}[theorem]{Definition}

\theoremstyle{remark}

\renewcommand{\vec}[1]{\boldsymbol{#1}}

\newcommand{\proposed}{{CompFS}}

\newcommand{\xv}{{\mathbf x}}
\newcommand{\bv}{{\mathbf b}}
\newcommand{\mv}{{\mathbf m}}
\newcommand{\zv}{{\mathbf z}}
\newcommand{\wv}{{\mathbf w}}
\newcommand{\Xv}{{\mathbf X}}
\newcommand{\Mv}{{\mathbf M}}
\newcommand{\Wv}{{\mathbf W}}

\newcommand{\Xc}{{\mathcal X}}
\newcommand{\Yc}{{\mathcal Y}}
\newcommand{\Fc}{{\mathcal F}}
\newcommand{\Sc}{{\mathcal S}}

\newcommand{\Gc}{{\mathcal G}}

\newcommand{\Zc}{{\mathcal Z}}

\newcommand{\E}{\mathbb{E}}

\title{Composite Feature Selection Using Deep Ensembles}

\author{%
  Fergus Imrie\thanks{Equal contribution}\\
  University of California, Los Angeles\\
  \texttt{imrie@ucla.edu}\\
  \And
  Alexander Norcliffe\footnotemark[1]\\
  University of Cambridge\\
  \texttt{alin2@cam.ac.uk}\\
  \And
  Pietro Li\`{o}\\
  University of Cambridge\\
  \texttt{pl219@cam.ac.uk}\\
  \And
  Mihaela van der Schaar\\
  University of Cambridge\\
  The Alan Turing Institute\\
  University of California, Los Angeles\\
  \texttt{mv472@cam.ac.uk}\\
}

\begin{document}

\maketitle
\setcounter{footnote}{0} 

\begin{abstract}
In many real world problems, features do not act alone but in combination with each other. For example, in genomics, diseases might not be caused by any single mutation but require the presence of multiple mutations. Prior work on feature selection either seeks to identify individual features or can only determine relevant groups from a predefined set. We investigate the problem of discovering groups of predictive features without predefined grouping. To do so, we define predictive groups in terms of linear and non-linear interactions between features. We introduce a novel deep learning architecture that uses an ensemble of feature selection models to find predictive groups, without requiring candidate groups to be provided. The selected groups are sparse and exhibit minimum overlap. Furthermore, we propose a new metric to measure similarity between discovered groups and the ground truth. We demonstrate the utility of our model on multiple synthetic tasks and semi-synthetic chemistry datasets, where the ground truth structure is known, as well as an image dataset and a real-world cancer dataset.
\end{abstract}

\section{Introduction}

Feature selection is a key problem permeating statistics, machine learning and broader science. Typically in high-dimensional datasets, the majority of features will not be responsible for the target response and thus an important goal is to identify which variables are truly predictive. For example, in healthcare there may be many features (such as age, sex, medical history, etc.) that could be considered, while only a small subset might in fact be relevant for predicting the likelihood of developing a specific disease. By eliminating irrelevant variables, feature selection algorithms can be used to drive discovery, improve model generalisation/robustness, and improve interpretability \cite{Chandrashekar2014survey}.

However, features often do not act alone but instead in \textit{combination}. 
In genetics, for instance, it has been noted that understanding the origins of many diseases may require methods able to identify more complex genetic models than single variants \cite{Papadimitriou2019predicting}.
More generally, often there are multiple groups of variables which act (somewhat) independently of each other. For example, in medicine or biology, a number of diseases can manifest from different mechanisms or pathways. Examples include cancer \cite{Meira2001}, amyotrophic lateral sclerosis (ALS) and frontotemporal dementia (FTD) \cite{Balendra2018}, inflammatory bowel disease \cite{Graham2020}, cardiovascular disease \cite{Kelly2009multiple}, and diabetes \cite{Merino2022}.

While feature selection might be able to identify a set of features associated with a particular response, the underlying structure of how features interact is not captured. Further, the resulting predictive models can be complex, hard to interpret, and not amenable to the generation of hypotheses that can be experimentally tested \cite{Knijnenburg2016logic}. This limits the impact such models can have in furthering scientific understanding across many domains where variables are known to interact, such as genetics \cite{Papadimitriou2019predicting,Phillips2008,Mani2008}, medicine \cite{Wilson2001,Brown2009}, and economics \cite{Balli2013}. 

Group feature selection is a generalisation of standard feature selection, where instead of selecting individual features, groups of features are either entirely chosen or entirely excluded. A primary application of group feature selection is when features are jointly measured, for example by different instruments. 
In such scenarios, groups are readily defined as features measured by the same instrument. A natural question is which instruments give the most meaningful measurements. Group feature selection has also been applied in situations where there is extensive domain knowledge regarding the group structure \cite{Rapaport2008classification} or where groups are defined by the correlation structure between features (e.g. neighbouring pixels in images are highly correlated). The pervasive issue with current group feature selection methods is that a predetermined grouping \emph{must} be provided, and the groups are selected from the given candidates. In reality, we may not know how to group the variables.

In this paper we seek to solve a related but ultimately different and more challenging problem, which we call \textit{Composite Feature Selection}. We wish to find groups of variables \textit{without} prior knowledge, where each group acts as a separate predictive subset of the features and the overall predictive power is greatest when all groups are used in unison.
We call each group of features a composite feature.\footnote{We will often refer to composite features as groups for brevity; in this paper, they refer to the same thing.}
By imposing this structure on the discovered features, we attempt to isolate pathways from features to the response variable. 
Discovering \textit{groups} of features offers deeper insights into \textit{why} specific features are important than standard feature selection.

\textbf{Contributions.}
(1) We formalise \textit{composite} feature selection as an extension of standard feature selection, defining composite features in terms of linear and non-linear interactions between variables (Sec. \ref{sec:problem}).
(2) We propose a new deep learning architecture for composite feature selection using an ensemble-based approach (Sec. \ref{sec:method}). 
(3) To assess our solution, we introduce a metric for assessing composite feature similarity based on Jaccard similarity (Sec. \ref{sec:exp}).
(4) We demonstrate the utility of our model on a range of synthetic and semi-synthetic tasks where the ground truth group features are known (Sec. \ref{sec:exp}). We see that our model not only frequently recovers the relevant features, but also often discovers the underlying group structure. We further illustrate our approach on an image dataset and a real-world cancer dataset, corroborating discovered features and feature interactions in the scientific literature.

\section{Related Work}

Significant attention has been placed on feature selection with a range of solutions including traditional methods (e.g. \cite{Liu1996filter,Kohavi1997wrapper,He2005laplacian}) and deep learning approaches \cite{Liang18bayesian,Yamada2020feature, abid2019concrete,lee2021self} (see Appendix \ref{app:ext_related_work} for further discussion of standard feature selection).
Several approaches have been extended to select predefined groups of variables instead of individual features.
For example, LASSO \cite{santosa1986linear, tibshirani1996regression} is a linear method that uses an L1 penalty to impose sparsity among coefficients. Group LASSO \cite{yuan2006model} generalises this to allow predefined groups to be selected or excluded jointly, rather than single features, by replacing the L1 penalty with L2 penalties on each group. Other feature selection methods, such as SLOPE \cite{bogdan2013statistical}, have been similarly extended to group feature selection to give Group-SLOPE \cite{brzyski2019group}. 
Further examples of group feature selection using adapted loss functions are SCAD-L2 \cite{zeng2014group} and hierarchical LASSO \cite{zhou2010group}. 
Similarly, Bayesian approaches to feature selection \cite{george1997approaches} have also been generalised to the group setting \cite{hernandez2013generalized}. Finally, the Knockoff procedure \cite{barber2015controlling, candes2018panning, jordon2018knockoffgan, liu2019deep, romano2020deep, sudarshan2020deep} is a generative procedure that creates fake covariates (knockoffs), obeying certain symmetries under permutations of real and knockoff features. By subsequently carrying out feature selection on the combined real and knockoff data, it is possible to obtain guarantees on the False Discovery Rate of the selected features. Generalisations of the Knockoff procedure to the group setting also exist \cite{dai2016knockoff, zhu2021deep}, where symmetries under permutations of entire groups must exist.

The key commonality is that none of these methods \textit{discover} groups, but instead can only \textit{select} groups from a set of predefined candidates. Therefore, while they may be applicable when we can split inputs into groups, they are not able to find groups of predictors on their own. Our work differs from these methods by considering the challenge of finding such groups in the absence of prior knowledge. Additionally, unlike prior work, we do not make assumptions about correlations between features or place restrictions on groups, such as requiring the candidate groups to partition the features.

\section{Problem Description}\label{sec:problem}

Let $\Xv \in \mathcal{X}^{p}$ be a $p$-dimensional signal (such as gene expressions or patient covariates) and $Y \in \mathcal{Y}$ be a response (such as disease traits). Informally, we wish to group features into the maximum number of subsets, $\Gc_i \subset [p]$, where the predictive power of any single group significantly decreases when any feature is removed, allowing us to separate the groups into different pathways from the signal to the response. Note that we do not enforce assumptions on the groups such as non-overlapping groups or every feature being in at least one group. 
In this section, we begin with a description of traditional feature selection before formalizing composite feature selection.

\subsection{Feature Selection}

The goal of traditional feature selection is to select a subset, $\Sc \subset [p]$, of features that are relevant for predicting the response variable. 
In particular, in the case of embedded feature selection \cite{Guyon2003introduction}, this is conducted jointly with the model selection process.

Let $*$ denote any point not in $\Xc$ and define $\Xc_{\Sc} = (\Xc \cup \{*\})^{p}$. 
Then, given $\Xv \in \Xc^{p}$, the selected subset of features can be denoted as $\Xv_{\Sc} \in \Xc_{\Sc}$ where $x_{\Sc,k}=x_{k}$ if $k\in\Sc$ and $x_{\Sc,k}=*$ if $k \notin \Sc$. 
Let $f: \Xc_{\Sc} \rightarrow \Yc$ be a function in some space $\Fc$ (such as the space of neural networks) taking subset $\Xv_{\Sc}$ as input to yield $Y$. Then, selecting relevant features for predicting a response can be achieved by solving the following optimization problem:
\begin{equation} \label{eq:objective_general}
    \underset{f\in \Fc,~\Sc \subset [p]}{\text{minimize}}~~~ \E_{
    \xv,y \sim p_{XY}}\Big[ \ell_{Y}\big(y, f(\xv_{\Sc}) \big)  \Big] ~~ \text{subject to}~~ |\Sc| \leq \delta,
\end{equation}
where $\delta$ constrains the number of selected features and $\ell_{Y}(y,y')$ is a task-specific loss function.

This can be solved by introducing a selection vector $\Mv = (M_{1},\cdots, M_{p}) \in \{0,1\}^{p}$, consisting of binary random variables governed by distribution $p_{M}$, with realization $\mv$ indicating selection of the corresponding features. 
Then, the selected features given vector $\mv$ can be written as
\begin{equation} \label{eq:selected_features}
    \tilde{\xv} \triangleq \mv \odot \xv + (1-\mv)\odot \hat{\xv},
\end{equation}
where $\odot$ indicates element-wise multiplication and $\hat{\xv}$ are the values assigned to features that are not selected (typically $\hat{\xv} \equiv 0$ or $\bar{\xv}$).
Eq. \eqref{eq:objective_general} can be (approximately) solved by jointly learning the model $f$ and the selection vector distribution $p_{M}$ based on the following optimization problem:
\begin{equation} \label{eq:objective_baseline}
  \underset{f,~p_{M}}{\text{minimize}}~~\E_{\xv,y \sim p_{XY}} \E_{\mv \sim p_{M}} \Big[ \ell_{Y}\big(y, f(\tilde{
   \xv}) \big)  + \beta \|\mv\|_{0} \Big],
\end{equation}
where $\beta$ is a balancing coefficient that controls the number of features to be selected. 

\subsection{Composite Feature Selection}

The goal of composite feature selection is not only to find the predictive features, but also to group them based on \textit{how} they are predictive. 
For example, assume features $x_1$ and $x_2$ are only predictive when both are known by the model, but have the same influence on the outcome independent of $x_3$. Then we wish to group $x_1, x_2$ separately from $x_3$. In this section, we define the embedded composite feature selection problem; that is, we want to find a valid model $f$ and groups $\{ \Gc_1 , \dots, \Gc_N \}$ in parallel. 
A model is only valid when the group representations are combined in a way where we can view each group as contributing an independent piece of information for the final prediction. A valid model acts on a \emph{set} of groups \cite{zaheer2017deep}, thus when combining groups, we require order not to matter. Therefore, we must combine the representations using a permutation invariant aggregator.

Let $A : ( \prod_{i} \mathbb{R}^{n} ) \xrightarrow{} \mathbb{R}^{N}$ be a general permutation invariant aggregation function. It is well established that for a specific choice of $\phi: \mathbb{R}^{n} \xrightarrow{} \mathbb{R}^{m}$ and $\rho: \mathbb{R}^{m} \xrightarrow{} \mathbb{R}^{N}$, $A$ can be decomposed as $\rho(\sum_i \phi(\cdot))$ (see \cite{zaheer2017deep} for examples). This gives $f(\xv) = g \big(\rho \big( \sum_{i} \phi(f_i(\xv_{\Gc_i})) \big)\big)$, where $f_i$ encodes group $i$, $\rho$ and $\phi$ give the permutation invariant aggregation, and $g$ is any final non-linear function, for instance softmax. The function composition of $\phi$ and $f_i$ can be relabelled as $\tilde{f}_i = \phi \circ f_i$, and the composition of $g$ and $\rho$ can be relabelled as $\tilde{\rho} = g\circ\rho$. This leads to $f(\xv) = \tilde{\rho} \big( \sum_{i} \tilde{f}_i(\xv_{\Gc_i}) \big)$, giving the following definition for a valid model structure in composite feature selection.

\begin{definition}
\label{def:group_feature_model}
The most general valid model for acting on $N$ composite features is given by:
\begin{equation}\label{group_def}
    f(\xv) = \rho 
    \bigg(
    \sum_{i=1}^{N}f_i(\xv_{\Gc_i})
    \bigg).
\end{equation}
That is, the groups must interact exactly once, all groups must be included, and the interaction is a summation; all other interactions can (and often should) be non-linear.
\end{definition}\vspace{-1mm}

Depending on the task, a specific permutation invariant aggregation may be chosen (e.g. \texttt{Max()}). However, any permutation invariant aggregator can be (approximately) expressed in the form of Def. \ref{def:group_feature_model}; thus, when learning from data, the general structure of Def. \ref{def:group_feature_model} means that this is not necessary.

The embedded composite feature selection problem can now be phrased in an analogous way to traditional feature selection.
Let $*$ denote some point not in $\Xc$ and define $\Xc_{\Gc_i} = (\Xc \cup \{*\})^{p}$. 
Then, given $\Xv \in \Xc^{p}$, the selected group of features is denoted as $\Xv_{\Gc_i} \in \Xc_{\Gc_i}$ where $x_{\Gc_i,k}=x_{k}$ if $k\in\Gc_i$ and $x_{k}=*$ if $k \notin \Gc_i$. 
Let $f_i: \Xc_{\Gc_i} \rightarrow \Zc$ be a function in $\Fc$ that takes as input the subset $\Xv_{\Gc_i}$ and outputs a latent representation $\zv_i$. Then, finding the groups of features can be achieved by solving the optimization problem:
\begin{equation} \label{eq:objective_group}
\begin{split}
    \underset{\rho, f_{i}\in \Fc,~ \Gc_{i} \subset [p]}{\text{minimize}}~~~ 
    \E_{
    \xv,y \sim p_{XY}}
    &
    \Bigg[ \ell_{Y}\bigg(y, 
    \rho
    \big(
    \sum_{i=1}^{N} f_i(\xv_{\Gc_i})
    \big)
    \bigg)  \Bigg]
\end{split}
\quad \quad
\begin{split}
    \text{subject to}~~ 
    \begin{aligned}
    &|\Gc_i| \leq \delta_i ~~ \forall{i},
    \\
    &
    N \geq \Delta,
    \end{aligned}
\end{split}
\end{equation}
where $\vec \delta$ constrains the number of selected features in each group and $\Delta$ gives the minimum number of groups. This objective leads to multiple smaller groups, rather than one group containing all features, which is consistent with our motivation of the problem. 

Continuing to expand from traditional feature selection, we can also extend the solution to the composite setting. For $N$ groups we can introduce a selection \emph{matrix} $\Mv \in \{0, 1 \}^{N \times p}$, governed by distribution $p_M$. For a realization $\Mv$, the selected features from group $i$ are given by
\begin{equation}
    \tilde{\xv}_i \triangleq \mv_i \odot \xv + (1-\mv_i)\odot \hat{\xv},
\end{equation}
where $\mv_i$ is the $i^{\text{th}}$ row of $\Mv$. We can approximately solve Eq. \eqref{eq:objective_group} by solving the optimization problem:
\begin{equation} \label{eq:objective_group_baseline}
  \underset{f,~p_{M}}{\text{minimize}}~~\E_{\xv,y \sim p_{XY}} \E_{\Mv \sim p_{M}} \Big[ \ell_{Y}\big(y, f(
   \xv) \big)  +  R_e(\Mv) \Big],
\end{equation}
where $f(\xv)$ obeys Def. \eqref{group_def} and $R_e$ is a regularisation term which controls how features are selected in each group. 
$R_e$ should capture both group size (i.e. encourage as few features as possible to be selected) but also the relationships between groups (i.e. groups should be distinct and not redundant).

\subsection{Challenges}
\label{sec:difficulty}
There are various challenges in solving the composite feature selection problem. 
While the ultimate task is to find predictive groups of features, there first remains the necessity simply to identify predictive features, which is already an NP-hard problem \cite{Amaldi1998approximability}.
Composite feature selection not only inherits this property but introduces additional complexity since we can think of each group as solving a separate feature selection problem. 
Consider the number of potential solutions: in traditional feature selection (assuming not all features are selected), there are $2^{n}-2$ ways of selecting a subset from $n$ features; even restricting to at most $m << n$ quickly becomes unfeasible for even modest values of $m$. 
In composite feature selection, \textit{every group} has the same number of solutions as traditional feature selection, drastically increasing the total number of possible solutions. A challenge specific to composite feature selection arises when the ground truth group structure contains groups with overlapping features (e.g. feature $x_1$ interacts independently with both $x_2$ and $x_3$). 
In this scenario, it is difficult to separate these two effects while penalizing the inclusion of additional features.

\begin{figure*}[ht]
\begin{center}
\centerline{\includegraphics[width=0.875\linewidth]{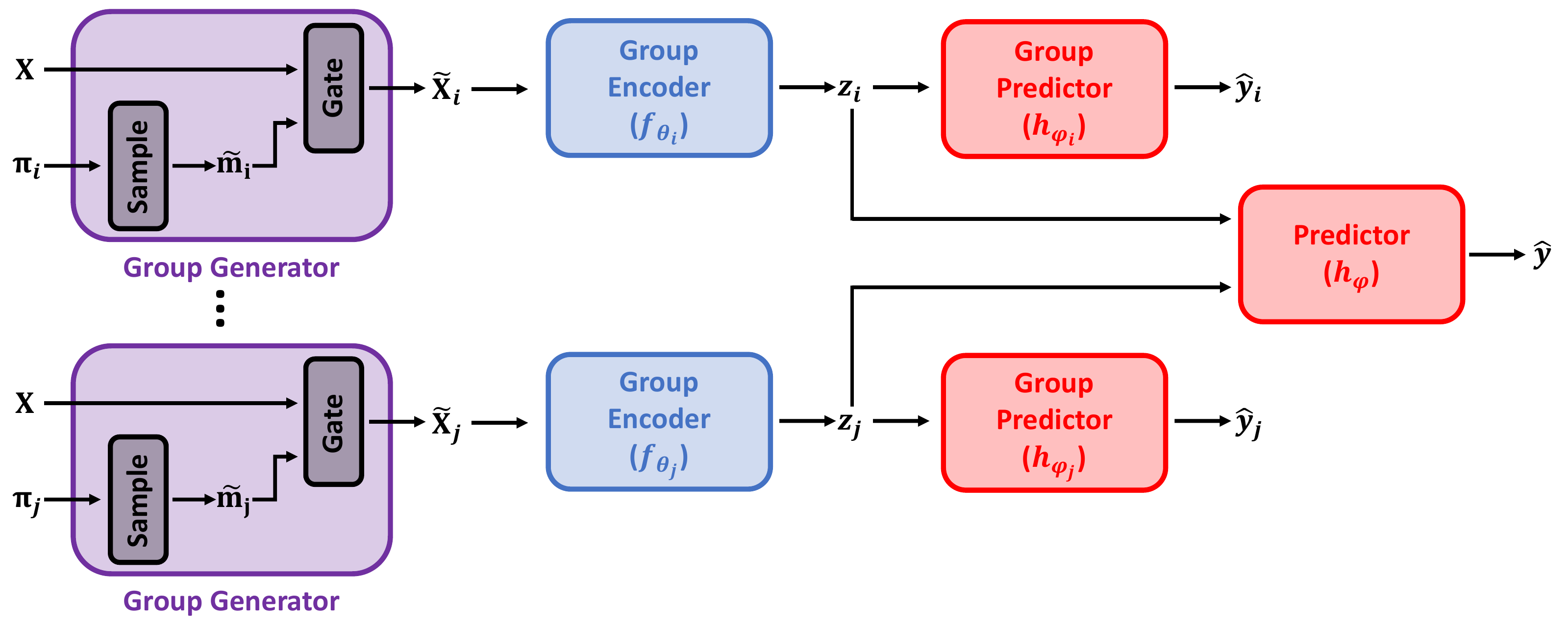}}
\vskip -0.1in
\caption{An illustration of {\proposed}. We use an ensemble of group selection models to discover composite features and an aggregate predictor to combine these features when issuing predictions.}
\label{fig:method}
\end{center}
\vskip -0.25in
\end{figure*}

\section{Method: {\proposed}}\label{sec:method}
In this section, we propose a novel architecture for finding predictive groups of features, which we refer to as \textbf{Comp}osite \textbf{F}eature \textbf{S}election ({\proposed}). In order to discover groups of features, our model is composed of a set of group selection models and an aggregate predictor. 
Our approach resembles an ensemble of ``weak'' feature selection models, where each learner attempts to solve the task using a sparse set of features (Figure \ref{fig:method}).
These models are then trained in such a way as to discover distinct predictive groups.
We first consider the group selection models in more detail before describing how the group selection models are combined and the training procedure.

\subsection{Group Selection Models}

{\proposed} is composed of a set of group selection models, each of which primarily aims to solve the traditional feature selection problem specified by Eq. \eqref{eq:objective_general}.
We achieve this by solving Eq. \eqref{eq:objective_baseline} using a neural network-based approach with stochastic gating of the input features. 
Each group selection model consists of the following three components (Figure \ref{fig:method}):
\begin{itemize}[leftmargin=5.0mm]\vspace{-1.0mm}
    \item \textit{Group Selection Probability}, $\bm{\pi}_i = (\pi_{1,i},\cdots,\pi_{p,i}) \in [0,1]^{p}$, which is a trainable vector that governs the Bernoulli distribution used to generate the gate vector $\mv_i$. Each element of the selection probability $\pi_{k,i}$ indicates the importance of the corresponding feature to the target.
    \item \textit{Group Encoder}, $f_{\theta_i}: \Xc^{p} \rightarrow \Zc$, that takes as input the selected subset of features $\tilde{\xv}_i$ and outputs latent representations $\zv_i \in \Zc$.
    \item \textit{Group Predictor}, $h_{\phi_i}: \Zc \rightarrow \Yc $, that takes as input the latent representations of the selected subset of features, $\zv_i=f_{\theta_i}(\tilde{\xv}_i)$, and outputs predictions on the target outcome.
\end{itemize}  \vspace{-1.0mm}

Solving Eq. \eqref{eq:objective_baseline} directly is not possible since the sampling step has no differentiable inverse.
Instead, we use the relaxed Bernoulli distribution \cite{Maddison17concrete,Jang17categorical} and apply the reparameterization trick as follows.

Formally, given selection probability $\bm{\pi} = (\pi_{1},\cdots, \pi_{p})$ and independent $\text{Uniform}(0,1)$ random variables $(U_{1}, \cdots, U_{p})$, we can generate a relaxed gate vector $\tilde{\mv}=(\tilde{m}_{1}, \cdots, \tilde{m}_{p}) \in (0,1)^{p}$ based on the following reparameterization trick \cite{Maddison17concrete}:
\begin{align} \label{eq:gate_generation}
\begin{split}
    \tilde{m}_{k} = \sigma \Big( \frac{1}{\tau} \big( &\log \pi_{k} - \log (1 - \pi_{k}) + \log U_{k} - \log (1 - U_{k}) \big) \Big),  
\end{split}
\end{align}
where $\sigma(x) = (1+\exp(-x))^{-1}$ is the sigmoid function.
This relaxation is parameterized by $\bm{\pi}$ and temperature $\tau \in (0, \infty)$.
Further, as $\tau \rightarrow 0$, the gate vectors $\tilde{m}_{k}$ converge to $\text{Bernoulli}(\pi_{k})$ random variables.
Crucially, this is differentiable with respect to $\bm{\pi}$.

Given group selection probability $\bm{\pi}_i$, we first sample relaxed Bernoulli random variable $\tilde{\mv}_i$ according to Eq. \eqref{eq:gate_generation} and then use $\tilde{\mv}_i$ in a gating procedure to select the group of features.
The output of the gate is:
\begin{equation} \label{eq:gate_output}
\tilde{\xv}_i = \text{gate}_i(\xv) = \tilde{\mv}_i \odot \xv + (1-\tilde{\mv}_i) \odot \bar{\xv},
\end{equation}
where we replace the variables that were not selected by their mean value $\bar{\xv}$. The mean is used because in certain tasks a feature having a value of 0 may be particularly meaningful. However, any (arbitrary) value could be used for non-selected features.
The gate output $\tilde{\xv}_i$ is then fed into the group encoder $f_{\theta_i}$ to yield representation 
$\zv_i = f_{\theta_i}(\tilde{\xv}_i).$
This representation is finally passed to the group predictor $h_{\phi_i}$ to produce the prediction for an individual learner,
$\hat{y}_i = h_{\phi_i}(\zv_i)$.

\subsection{Group Aggregation}

The final component necessary for {\proposed} is a way to aggregate the individual group selection models.
This is achieved via an overall \textit{predictor}, $h_{\phi}: \Zc \rightarrow \Yc $, that takes as input the set of latent representations $\{\zv_1,\dots, \zv_N\}$ produced by the individual learners and outputs predictions on the target outcome.
For simplicity, we apply a linear prediction head to the latent representations and use element-wise summation to aggregate.
Thus, the prediction of the ensemble is given by:
\begin{equation}
\hat{y} = h_{\phi}(\{\zv_1, \dots, \zv_N\}) = \rho \bigg[ \sum_{i=1}^N \Wv_i\zv_i + \bv_i \bigg], 
\end{equation}
where $N$ is the number of members of the ensemble (i.e. the number of groups) and $\rho$ is a suitable transformation (e.g. softmax). Note that by using element-wise summation, our model satisfies Def. \eqref{group_def} for acting on composite features.

\subsection{Loss Functions}

\paragraph{Group Selection Models.}
The individual learners can be trained to perform (traditional) feature selection (Eq. \eqref{eq:objective_general}) by minimizing the following loss function:
\begin{equation} \label{eq:loss_individual} 
    \mathcal{L}_{\Gc_i} = \E_{\xv,y\sim p_{XY}} \bigg[ \ell_{Y}\big(y, h_{\phi_i}(f_{\theta_i}( \text{gate}_i(\xv))) \big) + \beta \langle \bm{\pi}_{i} \rangle^2 
    \bigg],
\end{equation}
where $\ell_{Y}$ is a suitable loss function for the prediction task (e.g. cross-entropy for classification tasks and MSE for regression tasks) and $\beta \geq 0$ balances the two terms. 
Note the selections probabilities $\bm{\pi}_{i}$ are not regularized with the typical L1 penalty. Instead, we apply an L2 penalty to the mean selection probability $\langle \bm{\pi}_i \rangle$ for each individual learner. This is justified as follows.
Recall the optimization problem given by Eq. \eqref{eq:objective_group}. We desire a solution with the maximal number of predictive groups $N$ while minimizing the number of selected features per group $\sum_{i=1}^N | \mathcal{G}_i |$.
The standard L1 penalty term does not achieve this goal since adding an additional feature to either group $\mathcal{G}_i$ or $\mathcal{G}_j$ incurs the same penalty. In contrast, the L2 penalty imposed on $\langle \bm{\pi}_i \rangle$ penalizes adding extra features to already large groups, favoring the construction of smaller groups over larger ones.

\paragraph{Aggregate Predictor.}
The aggregate predictor can be trained jointly with the group selection models by minimizing a standard prediction loss (where $\ell_{Y}$ is the same as in Eq. \eqref{eq:loss_individual}):
\begin{equation} \label{eq:loss_aggregate_predictor} 
    \mathcal{L}_E = \E_{\xv,y\sim p_{XY}} \bigg[ \ell_{Y}\big(y, h_{\phi}(\{\zv_1, \dots, \zv_n\}) \big) \bigg].
\end{equation}

\paragraph{Additional Regularization.}
If we simply apply the losses given by Eqs. \eqref{eq:loss_individual}, \eqref{eq:loss_aggregate_predictor}, there will be limited (or even no) differentiation among the individual learners and the optimal solution would be for each learner to simply solve the traditional feature selection problem (Eq. \eqref{eq:objective_general}).
This results in all learners selecting the same features, which does not achieve our aim of discovering groups of predictive features.
In order to encourage differentiation between the models, we introduce an additional loss that penalizes the selection of the same features in multiple groups:
\begin{equation} \label{eq:loss_inter_group} 
    \mathcal{L}_R = \E_{\xv,y\sim p_{XY}} \bigg[ \sum_{i=1}^{N} \sum_{j>i} \bm{\pi}_{i} \cdot \bm{\pi}_{j} \bigg].
\end{equation}
\paragraph{Overall Loss.}
Combining the above, our overall loss function therefore can be written as follows:
\begin{equation} \label{eq:loss_overall} 
    \mathcal{L} = \sum_{i=1}^N \mathcal{L}_{\Gc_i} + \beta_E \mathcal{L}_E + \beta_R \mathcal{L}_R,
\end{equation}
where $\beta_E, \beta_R \geq 0$ are hyperparameters to balance the losses.

Training {\proposed} with the loss given by Eq. \eqref{eq:loss_overall} is designed to achieve the following: (1) The overall ensemble network should be a good predictor ($\mathcal{L}_E$). (2) Each individual learner should solve the traditional feature selection problem ($\mathcal{L}_{\Gc_i}$), which requires the group predictor to be accurate while selecting minimal features. 
However, the individual learners should not be maximally predictive by definition (hence why we compare individual group feature selection models to weak learners). 
(3) Finally, we want the groups to be distinct and thus discourage highly similar groups ($\mathcal{L}_R$). However, note that we do not exclude the possibility of some overlap of features between groups. 
The model is end-to-end differentiable, so we train with gradient descent. 

\textbf{Evaluation.}
During evaluation, only the gating procedure changes. The way features can be selected is chosen by the user. A standard solution which we adopt in this paper is using a threshold $\lambda$ and computing gate vectors $\mv_i$ as follows:
$m_{i,k} = 1, \text{if}\ \pi_{i,k} > \lambda$ and $0$ otherwise.

\section{Experiments}\label{sec:exp}

We evaluate {\proposed} using several synthetic and semi-synthetic datasets where ground truth feature importances and group structure are known. 
In addition, we illustrate our method on an image dataset (MNIST) and a real-world cancer dataset (METABRIC). 
Specific architectural details are given in App. \ref{app:architecture}. Additional information regarding experiments, benchmarks, and datasets can be found in App. \ref{app:experiments}.
Additional ablations and sensitivity analysis are in App. \ref{app:ablation}.
The code for our method and experiments is available on Github.
\footnote{\url{https://github.com/a-norcliffe/Composite-Feature-Selection}}
\footnote{\url{https://github.com/vanderschaarlab/Composite-Feature-Selection}}

\vspace{-3mm}

\paragraph{Benchmarks.}
The primary goal of our experiments is to demonstrate the utility of discovering composite features over traditional feature selection. Our main benchmark is an oracle feature selection method (``Oracle") that perfectly selects the ground truth features but provides no structure, giving all features as one group.
By definition, this is the strongest standard feature selection baseline for the scenarios where the ground truth features are known. We also include comparisons to a linear feature selection method (LASSO) \cite{tibshirani1996regression} and two non-linear, state of the art approaches, Stochastic Gates (STG) \cite{Yamada2020feature} and Supervised Concrete Autoencoder (Sup-CAE) \cite{abid2019concrete}. Finally, we compare with Group LASSO \cite{yuan2006model}, where we enumerate all groups with 1 or 2 features as predefined groups. Note this represents a significant simplification of the task for Group Lasso. We include additional baselines in App. \ref{app:add_experiments}.

\vspace{-3mm}

\paragraph{Metrics.} 
When the ground truth feature groups $\Gc_1, \dots, \Gc_N$ are known, we use True Positive Rate (TPR) and False Discovery Rate (FDR) to assess the discovered features against the ground truth. To assess composite features, i.e. grouping, we define the Group Similarity (G\textsubscript{sim}) as the normalized Jaccard similarity between ground truth feature groups and the most similar proposed group: 
\begin{equation}
\text{G\textsubscript{sim}} = \frac{1}{\max(N,K)} \sum_{i=1}^{N} \max_{j \in [K]} \mathcal{J}(\Gc_i, \hat{\Gc}_j), 
\end{equation}
where $\mathcal{J}$ is the Jaccard index \cite{Jaccard1912} and $\hat{\Gc}_1, \dots, \hat{\Gc}_K$ are the discovered groups. G\textsubscript{sim}$\in [0,1]$, where G\textsubscript{sim}$=1$ corresponds to perfect recovery of the ground truth groups, while G\textsubscript{sim}$=0$ when none of the correct features are discovered (see App. \ref{app:jaccard} for additional details together with examples). We assess the models by seeing if the ground truth features have been correctly discovered, using TPR and FDR. We then see if the underlying grouping has been uncovered (and correct features) using G\textsubscript{sim}.
Finally, we assess the predictive power of the discovered features using accuracy or area under the receiver operating curve (AUROC).

\subsection{Synthetic Experiments.}

\paragraph{Dataset Description.}
We begin by evaluating our method on a range of synthetic datasets where the ground truth feature importance is known (Table \ref{Table:Gauss}). We generate synthetic datasets by sampling from the Gaussian distribution with initially no correlations among features ($X \sim \mathcal{N}(0,I)$). We construct binary classification tasks, where the class $y$ is determined by the following decision rules:

\begin{table*}[ht]
	\vskip -0.05in
	\caption{Performance on Synthetic Datasets, values are recorded with their standard deviations.} \label{Table:Gauss}
	\vskip -0.05in
	\begin{center}
    \begin{small}
    \begin{sc}
        \resizebox{0.9\linewidth}{!}{
		\begin{tabular}[b]{c c c c c c c}
			\toprule
			\textbf{Dataset} & \textbf{Model} & \textbf{TPR} &\textbf{FDR} & \textbf{G\textsubscript{sim}} 
			& \textbf{No. Groups} & \textbf{Accuracy} (\%)\\
			\midrule
			\multirow{6}{*}{Syn1} 
			& CompFS(5) & 100.0 {\scriptsize $\pm$ 0.0} & 0.0 {\scriptsize $\pm$ 0.0} &  0.91 {\scriptsize $\pm$ 0.14} & 2.2 {\scriptsize $\pm$ 0.4} &  98.9 {\scriptsize $\pm$ 0.5}   \\
			 & Oracle    & 100.0 {\scriptsize $\pm$ 0.0}  & 0.0 {\scriptsize $\pm$ 0.0}   &  0.50 {\scriptsize $\pm$ 0.00}  & 1.0 {\scriptsize $\pm$ 0.0} & 100.0 {\scriptsize $\pm$ 0.0} \\
			 & LASSO & 100.0 {\scriptsize $\pm$ 0.0}  & 0.0 {\scriptsize $\pm$ 0.0}   &  0.50 {\scriptsize $\pm$ 0.00}  & 1.0 {\scriptsize $\pm$ 0.0} & 81.8 {\scriptsize $\pm$ 2.0} \\
			 & Group LASSO & 100.0 {\scriptsize $\pm$ 0.0}  & 0.0 {\scriptsize $\pm$ 0.0}   &  0.67 {\scriptsize $\pm$ 0.00}  & 3.0 {\scriptsize $\pm$ 0.0} & 83.8 {\scriptsize $\pm$ 1.4} \\
			 & STG & 100.0 {\scriptsize $\pm$ 0.0} & 0.0 {\scriptsize $\pm$ 0.0} &  0.50 {\scriptsize $\pm$ 0.00} & 1.0 {\scriptsize $\pm$ 0.0} &  97.8 {\scriptsize $\pm$ 1.4}\\
			 & Sup-CAE & 100.0 {\scriptsize $\pm$ 0.0} & 0.0 {\scriptsize $\pm$ 0.0} &  0.50 {\scriptsize $\pm$ 0.00} & 1.0 {\scriptsize $\pm$ 0.0} &  97.8 {\scriptsize $\pm$ 1.4} \\
			\midrule
			\multirow{6}{*}{Syn2} 
			& CompFS(5)  &  95.0 {\scriptsize $\pm$ 15.0} & 0.0 {\scriptsize $\pm$ 0.0}  &   0.90 {\scriptsize $\pm$ 0.20} & 1.8 {\scriptsize $\pm$ 0.4}	& 95.5 {\scriptsize $\pm$ 5.4} \\
			 & Oracle    & 100.0 {\scriptsize $\pm$ 0.0}  & 0.0 {\scriptsize $\pm$ 0.0}  &  0.50 {\scriptsize $\pm$ 0.00} & 1.0 {\scriptsize $\pm$ 0.0} & 100.0 {\scriptsize $\pm$ 0.0} \\
			 & LASSO & 0.0 {\scriptsize $\pm$ 0.0}  & 0.0 {\scriptsize $\pm$ 0.0}  &  0.00 {\scriptsize $\pm$ 0.00} & 0.0 {\scriptsize $\pm$ 0.0} & 52.6 {\scriptsize $\pm$ 2.9} \\
			 & Group LASSO & 0.0 {\scriptsize $\pm$ 0.0}  & 0.0 {\scriptsize $\pm$ 0.0}   &  0.00 {\scriptsize $\pm$ 0.00}  & 0.0 {\scriptsize $\pm$ 0.0} & 52.2 {\scriptsize $\pm$ 0.9} \\
			 & STG & 100.0 {\scriptsize $\pm$ 0.0} & 0.0 {\scriptsize $\pm$ 0.0} &  0.50 {\scriptsize $\pm$ 0.00} & 1.0 {\scriptsize $\pm$ 0.0} &  93.9 {\scriptsize $\pm$ 2.2}\\
			 & Sup-CAE & 37.5 {\scriptsize $\pm$ 31.7} & 42.5 {\scriptsize $\pm$ 44.2} &  0.24 {\scriptsize $\pm$ 0.20} & 1.0 {\scriptsize $\pm$ 0.0} &  61.9 {\scriptsize $\pm$ 12.8}\\
			\midrule
			\multirow{6}{*}{Syn3} 
			& CompFS(5) & 100.0 {\scriptsize $\pm$ 0.0} & 0.0 {\scriptsize $\pm$ 0.0}  & 0.68 {\scriptsize $\pm$ 0.05} & 1.3 {\scriptsize $\pm$ 0.5} & 97.4 {\scriptsize $\pm$ 1.1} \\
			 & Oracle    & 100.0 {\scriptsize $\pm$ 0.0} & 0.0 {\scriptsize $\pm$ 0.0}  &  0.67 {\scriptsize $\pm$ 0.00}  & 1.0 {\scriptsize $\pm$ 0.0}    & 100.0 {\scriptsize $\pm$ 0.0}    \\
			 & LASSO & 0.0 {\scriptsize $\pm$ 0.0}  & 0.0 {\scriptsize $\pm$ 0.0}  &  0.00 {\scriptsize $\pm$ 0.00} & 0.0 {\scriptsize $\pm$ 0.0} & 56.5 {\scriptsize $\pm$ 4.0} \\
			 & Group LASSO & 0.0 {\scriptsize $\pm$ 0.0}  & 0.0 {\scriptsize $\pm$ 0.0}   &  0.00 {\scriptsize $\pm$ 0.00}  & 0.0 {\scriptsize $\pm$ 0.0} & 54.6 {\scriptsize $\pm$ 1.3} \\
			 & STG & 100.0 {\scriptsize $\pm$ 0.0} & 0.0 {\scriptsize $\pm$ 0.0} &  0.67 {\scriptsize $\pm$ 0.00} & 1.0 {\scriptsize $\pm$ 0.0} &  95.3 {\scriptsize $\pm$ 1.7}\\
			 & Sup-CAE & 23.3 {\scriptsize $\pm$ 31.6} & 66.7 {\scriptsize $\pm$ 47.1} &  0.23 {\scriptsize $\pm$ 0.31} & 1.0 {\scriptsize $\pm$ 0.0} &  62.6 {\scriptsize $\pm$ 12.6}\\
			\midrule
			\multirow{6}{*}{Syn4} 
			& CompFS(5)   & 90.0 {\scriptsize $\pm$ 12.2} & 51.9 {\scriptsize $\pm$ 13.8} &  0.47 {\scriptsize $\pm$ 0.20} & 2.5 {\scriptsize $\pm$ 0.7} 			& 95.8 {\scriptsize $\pm$ 1.8} \\
			 & Oracle    & 100.0 {\scriptsize $\pm$ 0.0} & 0.0 {\scriptsize $\pm$ 0.0}  &  0.50 {\scriptsize $\pm$ 0.00} & 1.0 {\scriptsize $\pm$ 0.0} & 100.0 {\scriptsize $\pm$ 0.0} \\
			 & LASSO & 0.0 {\scriptsize $\pm$ 0.0}  & 0.0 {\scriptsize $\pm$ 0.0}  &  0.00 {\scriptsize $\pm$ 0.00} & 0.0 {\scriptsize $\pm$ 0.0} & 51.8 {\scriptsize $\pm$ 3.2} \\
			 & Group LASSO & 0.0 {\scriptsize $\pm$ 0.0}  & 10.0 {\scriptsize $\pm$ 31.6}   &  0.00 {\scriptsize $\pm$ 0.00}  & 0.1 {\scriptsize $\pm$ 0.3} & 53.0 {\scriptsize $\pm$ 1.1} \\
			 & STG & 100.0 {\scriptsize $\pm$ 0.0} & 66.7 {\scriptsize $\pm$ 0.0} &  0.17 {\scriptsize $\pm$ 0.00} & 1.0 {\scriptsize $\pm$ 0.0} &  94.2 {\scriptsize $\pm$ 2.1}\\
			 & Sup-CAE & 72.5 {\scriptsize $\pm$ 14.2} & 16.7 {\scriptsize $\pm$ 14.7} &  0.39 {\scriptsize $\pm$ 0.08} & 1.0 {\scriptsize $\pm$ 0.0} &  72.2 {\scriptsize $\pm$ 13.2}\\
			\bottomrule
        \end{tabular} 
        }
 	\end{sc}
 	\end{small}
 	\end{center}
    \hfill
    \vskip -0.1in
\end{table*}

\begin{itemize}[leftmargin=5.0mm]\vspace{-1.5mm}
\setlength{\itemsep}{0pt}
    \item \textbf{(Syn1)} $y=1$ if $x_1 > 0.55$ or $x_2 > 0.55$, 0 otherwise. The ground truth groups are $\{ \{ 1 \}, \{2\} \}$. This task assesses whether the model can separate two features rather than group them together.
    \item \textbf{(Syn2)} $y=1$ if $x_1x_2 > 0.30$ or $x_3x_4 > 0.30$, 0 otherwise. The ground truth groups are $\{ \{ 1, 2 \}, \{3, 4\} \}$. This task requires identifying groups consisting of more than one variable.
    \item \textbf{(Syn3)} $y=1$ if $x_1x_2 > 0.30$ or $x_1x_3 > 0.30$, 0 otherwise. The ground truth groups are $\{ \{ 1, 2 \}, \{1, 3\} \}$. This task investigates whether a model can split the features into two \emph{overlapping} groups of two, rather than one group with all three features.
    \item \textbf{(Syn4)} $y=1$ if $x_1x_4 > 0.30$ or $x_7x_{10} > 0.30$, 0 otherwise. The ground truth groups are $\{ \{ 1, 4 \}, \{7, 10\} \}$. This task is equivalent to \textbf{Syn2}, however, here the features exhibit strong correlation in collections of 3, i.e. features 1, 2, and 3 are highly correlated, features 4, 5, and 6 are highly correlated, and so on. This task demonstrates the difficulty of carrying out group feature selection (and indeed standard feature selection) when the features are highly correlated.
\end{itemize}

The decision rules are created such that there is minimal class imbalance. We use signals with 500 dimensions to demonstrate the utility in the high dimensional regime. We use 20,000 samples to train and 200 to test. Each experiment is repeated 10 times.

\vspace{-3mm}

\paragraph{Analysis.}
On both Syn1 and Syn2, {\proposed} achieves high TPR with no false discoveries (0\% FDR) and significantly higher G\textsubscript{sim} than the Oracle. 
Despite allowing {\proposed} to discover up to 5 groups, {\proposed} typically finds the correct number of groups (2), demonstrating that it is not necessary for the number of potential composite features to match the ground truth, which is vital in real-world use cases where this is unknown. 
Syn3 is significantly more challenging due to the overlapping structure and we observe essentially the same performance as Oracle. 
Despite finding all the correct features and no false discoveries, {\proposed} typically finds the union $\{ 1, 2, 3 \}$ rather than the underlying group structure $\{ \{ 1, 2 \}, \{1, 3\} \}$.
Finally, for Syn4, while {\proposed} has a relatively high FDR, it frequently finds the ground truth relevant features and groups with similar G\textsubscript{sim} to Oracle. This is a challenging task with significant correlation between features. Despite this, {\proposed} is able to uncover the underlying group structure, providing additional insight over traditional feature selection. 
STG typically performs reasonably in terms of traditional feature selection, but scores poorly in terms of G\textsubscript{sim} due to not providing any group information.

\subsection{Semi-Synthetic Experiments.}

\paragraph{Dataset Description.}
Next, we assess our ability to identify composite features using semi-synthetic molecular datasets.
These tasks are analogs of real-world problems, such as identifying biologically active chemical groups; however, the labels are determined by a synthetic ``binding logic'' so that the ground truth feature relevance is known. 
We use several of the datasets constructed by \cite{McCloskey2019attribution}, some of which were also used by \cite{Sanchez-Lengeling2020evaluating}.\footnote{Data from \url{https://github.com/google-research/graph-attribution/raw/main/data/all_16_logics_train_and_test.zip}.}
The synthetic ``binding logics'' are expressed as a combination of molecular fragments that must either be present or absent for binding to occur and are used to label molecules from the ZINC database \cite{Irwin2012zinc}. 
Each logic includes up to four functional groups (Table \ref{Table:chem_logics}).
Molecules are featurized using a set of 84 functional groups, where feature $x_i=1$ if the molecule contains functional group $i$ and $0$ otherwise. The specific binding logics are given in App. \ref{app:chem}.

\begin{table*}[ht]
	\vskip -0.05in
	\caption{Performance on Chemistry Datasets, values are recorded with their standard deviations.} \label{Table:CHEM}
	\vskip -0.05in
	\begin{center}
    \begin{small}
    \begin{sc}
        \resizebox{0.9\linewidth}{!}{
		\begin{tabular}[b]{c c c c c c c}
			\toprule
			\textbf{Dataset} & \textbf{Model} & \textbf{TPR} &\textbf{FDR} & \textbf{G\textsubscript{sim}} 
			& \textbf{No. Groups} & \textbf{Accuracy} (\%)\\
			\midrule
			\multirow{6}{*}{Chem1}  
			& CompFS(5)  &  100.0 {\scriptsize $\pm$ 0.0} & 0.0 {\scriptsize $\pm$ 0.0}  &  0.82 {\scriptsize $\pm$ 0.20} &  1.9 {\scriptsize $\pm$ 0.5} & 100.0 {\scriptsize $\pm$ 0.0} \\
			& Oracle    & 100.0 {\scriptsize $\pm$ 0.0}  & 0.0 {\scriptsize $\pm$ 0.0}  &  0.50 {\scriptsize $\pm$ 0.00} & 1.0 {\scriptsize $\pm$ 0.0} & 100.0 {\scriptsize $\pm$ 0.0} \\
			& LASSO & 100.0 {\scriptsize $\pm$ 0.0} & 0.0 {\scriptsize $\pm$ 0.0} & 0.50 {\scriptsize $\pm$ 0.00} & 1.0 {\scriptsize $\pm$ 0.0} & 75.8 {\scriptsize $\pm$ 0.0} \\
			& Group LASSO & 100.0 {\scriptsize $\pm$ 0.0}  & 0.0 {\scriptsize $\pm$ 0.0}   &  0.67 {\scriptsize $\pm$ 0.00}  & 3.0 {\scriptsize $\pm$ 0.0} & 100.0 {\scriptsize $\pm$ 0.0} \\
			& STG & 100.0 {\scriptsize $\pm$ 0.0} & 0.0 {\scriptsize $\pm$ 0.0} &  0.50 {\scriptsize $\pm$ 0.00} & 1.0 {\scriptsize $\pm$ 0.0} &  100.0 {\scriptsize $\pm$ 0.0}\\
			& Sup-CAE & 62.5 {\scriptsize $\pm$ 13.2} & 23.3 {\scriptsize $\pm$ 17.5} &  0.37 {\scriptsize $\pm$ 0.07} & 1.0 {\scriptsize $\pm$ 0.0} &  77.8 {\scriptsize $\pm$ 11.0}\\
			\midrule
			\multirow{6}{*}{Chem2}  
			& CompFS(5)   & 100.0 {\scriptsize $\pm$ 0.0}  &
			0.0 {\scriptsize $\pm$ 0.0} &  0.72 {\scriptsize $\pm$ 0.24} & 2.2 {\scriptsize $\pm$ 0.6}
			& 100.0 {\scriptsize $\pm$ 0.0} \\
			& Oracle    & 100.0 {\scriptsize $\pm$ 0.0} & 0.0 {\scriptsize $\pm$ 0.0}  &  0.50 {\scriptsize $\pm$ 0.00} & 1.0 {\scriptsize $\pm$ 0.0} & 100.0 {\scriptsize $\pm$ 0.0} \\
			& LASSO & 100.0 {\scriptsize $\pm$ 0.0} & 0.0 {\scriptsize $\pm$ 0.0} & 0.50 {\scriptsize $\pm$ 0.00} & 1.0 {\scriptsize $\pm$ 0.0} & 81.6 {\scriptsize $\pm$ 0.0} \\
			& Group LASSO & 100.0 {\scriptsize $\pm$ 0.0}  & 0.0 {\scriptsize $\pm$ 0.0}   &  0.40 {\scriptsize $\pm$ 0.00}  & 5.0 {\scriptsize $\pm$ 0.0} & 81.6 {\scriptsize $\pm$ 0.0} \\
			& STG & 100.0 {\scriptsize $\pm$ 0.0} & 0.0 {\scriptsize $\pm$ 0.0} &  0.50 {\scriptsize $\pm$ 0.00} & 1.0 {\scriptsize $\pm$ 0.0} &  100.0 {\scriptsize $\pm$ 0.0}\\
			& Sup-CAE & 66.7 {\scriptsize $\pm$ 0.0} & 0.0 {\scriptsize $\pm$ 0.0} &  0.42 {\scriptsize $\pm$ 0.00} & 1.0 {\scriptsize $\pm$ 0.0} &  80.9 {\scriptsize $\pm$ 9.5}\\
			 \midrule
			\multirow{6}{*}{Chem3} 
			& CompFS(5)   & 100.0 {\scriptsize $\pm$ 0.0} & 7.3 {\scriptsize $\pm$ 11.7} &  0.62 {\scriptsize $\pm$ 0.17} & 2.4  {\scriptsize $\pm$ 0.5}
			& 100.0 {\scriptsize $\pm$ 0.0} \\
			& Oracle    & 100.0 {\scriptsize $\pm$ 0.0} & 0.0 {\scriptsize $\pm$ 0.0}  &  0.50 {\scriptsize $\pm$ 0.00} & 1.0 {\scriptsize $\pm$ 0.0} & 100.0 {\scriptsize $\pm$ 0.0} \\
			& LASSO & 100.0 {\scriptsize $\pm$ 0.0} & 0.0 {\scriptsize $\pm$ 0.0} & 0.50 {\scriptsize $\pm$ 0.00} & 1.0 {\scriptsize $\pm$ 0.0} & 87.4 {\scriptsize $\pm$ 5.2} \\
			& Group LASSO & 100.0 {\scriptsize $\pm$ 0.0}  & 20.0 {\scriptsize $\pm$ 0.0}   &  0.20 {\scriptsize $\pm$ 0.00}  & 10.0 {\scriptsize $\pm$ 0.0} & 91.5 {\scriptsize $\pm$ 0.0} \\
			& STG & 100.0 {\scriptsize $\pm$ 0.0} & 0.0 {\scriptsize $\pm$ 0.0} &  0.50 {\scriptsize $\pm$ 0.00} & 1.0 {\scriptsize $\pm$ 0.0} &  100.0 {\scriptsize $\pm$ 0.0}\\
			& Sup-CAE & 62.5 {\scriptsize $\pm$ 13.2} & 23.3 {\scriptsize $\pm$ 17.5} &  0.37 {\scriptsize $\pm$ 0.07} & 1.0 {\scriptsize $\pm$ 0.0} &  77.8 {\scriptsize $\pm$ 11.0}\\
			\bottomrule
        \end{tabular} 
        }
 	\end{sc}
 	\end{small}
 	\end{center}
    \hfill
    \vskip -0.2in
\end{table*}

\paragraph{Analysis.}
All methods are able to identify the ground truth relevant features; however, only {\proposed} provides deeper insights.
Unlike for Syn1-4, LASSO correctly selects the ground truth features since the dataset consists of binary variables and thus it is possible to find performant linear models.  However, while discovering the correct features, Group LASSO selects all possible combinations of these features, adding no benefit over standard feature selection. 

For Chem1-2, {\proposed} perfectly recovers the group structure in the majority of experiments, leading to high G\textsubscript{sim} far exceeding traditional feature selection.
On Chem3, we occasionally discover additional features that are not part of the binding logic. 
However, a number of molecular fragments are strongly correlated with the binding logic, even though they are not themselves included.
In fact, some features contain information about \textit{multiple} functional groups.
For example, esters contain a carbonyl and an ether; both are in the binding logic for Chem3, while ester is not, despite being highly informative, and thus occasionally {\proposed} incorrectly selects this feature. In spite of this, {\proposed} achieves significantly higher G\textsubscript{sim} than even Oracle. This demonstrates the benefit of the grouping discovered by {\proposed}, even with a modest number of false discoveries. As before, {\proposed} typically finds the correct number of groups (2), despite being able to discover up to 5 groups, further demonstrating that the number of composite features need not be known \textit{a priori}, which is the case in real-world applications.

\subsection{MNIST}

\begin{wrapfigure}{r!}{0.55\textwidth}
\vspace{-0.55cm}
    \centering
    \includegraphics[width=\linewidth]{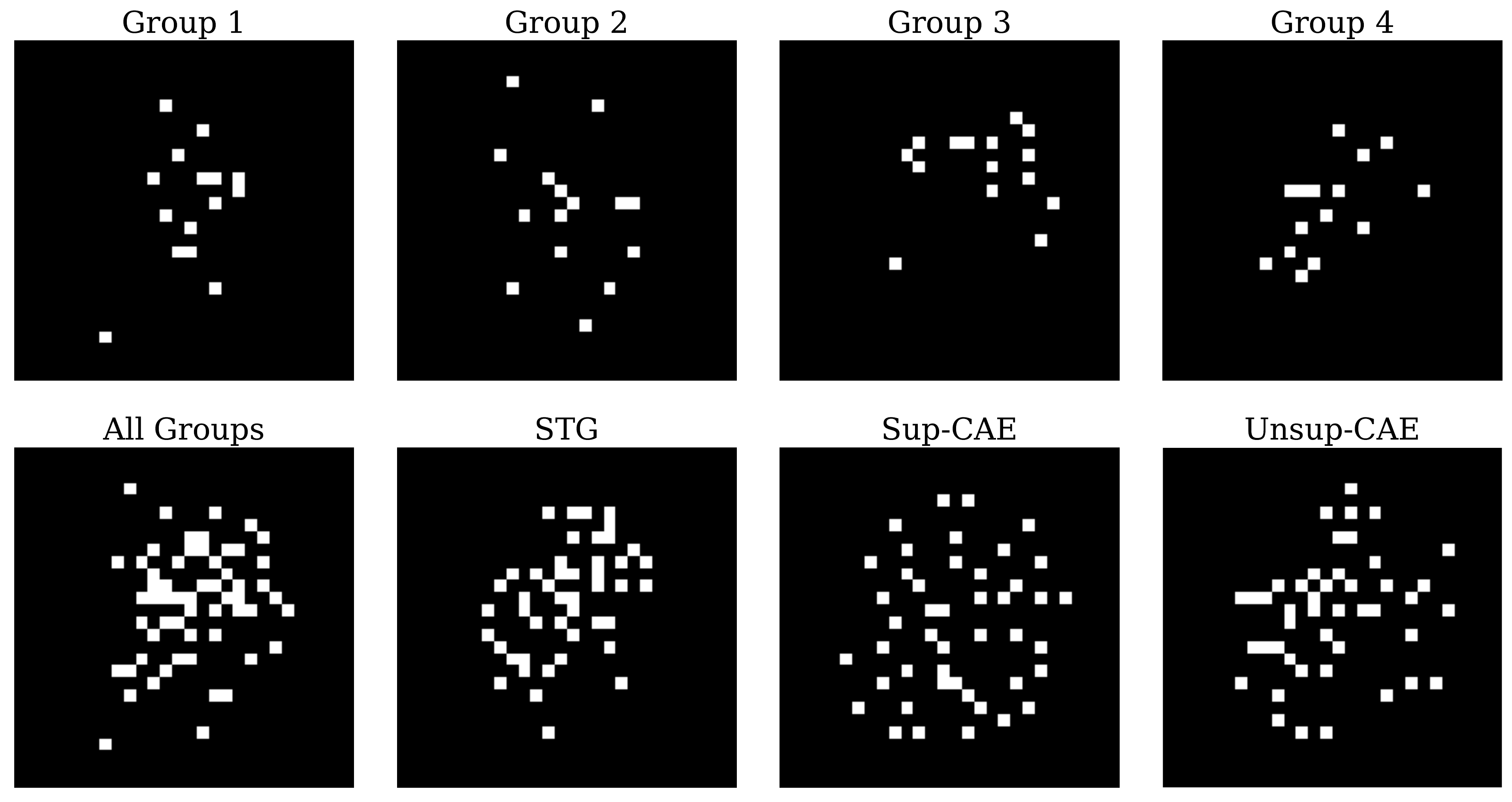}
    \caption{Pixels selected by {\proposed} and baselines.}
    \label{fig:mnist_discovered}
    \vspace{-2mm}
\vspace{-0.55cm}
\end{wrapfigure}
\paragraph{Dataset Description.} 
We investigate \proposed{} on the MNIST dataset \cite{lecun1998gradient}. While this well-known dataset consists of 28x28 images and typically fixed pixel locations do not have specific meaning, it has been extensively used in the feature selection literature due to the handwritten digits being centered and scaled, thus each of the 784 pixels can be (somewhat) meaningfully treated as a separate feature.  
While the ground truth group structure is unknown, a benefit of MNIST is that we can readily visualise selected features. 

\paragraph{Analysis.}
The features discovered by \proposed{} (using 4 groups), STG, Sup-CAE and Unsup-CAE are shown in Figure \ref{fig:mnist_discovered}. As expected, all selected pixels are central and relatively spread out. However, the four groups discovered by {\proposed} appear to have slightly different focus, in particular Group 3. 

To investigate the impact of these differences, we evaluate the predictive power of each of the groups. We find that the individual groups have relatively low accuracies (72\%-81\%), in part due to only using 15 pixels (<2\% of total). However, the union of these features achieve significantly greater accuracy of 95\%, equalling the performance of STG and (Sup-,Unsup-)CAE. This illustrates that the grouping does not seem to drastically affect performance, despite enforcing constraints on how the features are used by the model. 

Finally, we consider the per-class accuracy of each of these groups. Interestingly, the variance in performance between classes is significantly higher for the groups than any of the overall methods (Fig. \ref{fig:mnist_bar}). For example, Group 2 struggles to identify digits 4 and 5, while Group 3 performs poorly on digits 2, 3, and 8. This highlights the distinct information contained in each group.
Further details and analysis is provided in App. \ref{app:images}.

\subsection{Real-World Data: METABRIC}

\vskip -0.06in
\begin{wraptable}{r}{0.40\textwidth}
	\vskip -0.4in
	\caption{METABRIC performance. We compare CompFS and STG using 25 features to an MLP using all 489 features.} 
	\label{Table:METABRIC}
	\vskip +0.1in
	\centering
    \fontsize{8}{9}\selectfont
		\begin{tabular}[b]{c c}
			\toprule
			\textbf{Model} & \textbf{AUROC} \\
			\midrule
			MLP (All features) & 0.869 \\
			CompFS(5)          & 0.830  \\
			STG                & 0.843 \\
			\bottomrule
        \end{tabular} 
    \hfill
    \vskip -0.3in
\end{wraptable}

\paragraph{Dataset Description.} Finally, we assess CompFS on a real-world dataset, METABRIC \citep{Curtis2012,Pereira2016}, where the ground truth group structure is \textit{unknown}. METABRIC contains gene expression, mutation, and clinical data for 1,980 primary breast cancer samples. We evaluated the ability to predict the progesterone receptor (PR) status of the tissue based on the gene expression data, which consists of measurements for 489 genes. 

\vspace{-3mm}

\paragraph{Analysis.}
{\proposed} suffers limited performance degradation compared to using all features, despite only using 5\% of the features (Table \ref{Table:METABRIC}).
Despite imposing a more rigid structural form on how features can interact in the predictive model, STG only had marginally greater predictive power than {\proposed}. However, {\proposed} provides greater insight into how the features interact than STG.

We found supporting evidence in the scientific literature for all but 1 of the genes discovered by {\proposed} (Table \ref{Table:METABRIC_features}). In addition, within each group, we found further evidence of the interactions between genes, demonstrating the ability for CompFS to learn informative groups of features.
For example, in Group 1, CXCR1 and PEN-2 (the protein encoded by PSENEN) are known to interact \citep{Bakele2014}. In Group 2, BMP6 encodes a member of the TGF-$\beta$ superfamily of proteins, and TGF-$\beta$ triggers activation of SMAD3 \citep{Chen2022}. In the same group, MAPK1 activity is dependent on the activity of PRKCQ in breast cancer cells \citep{Byerly2016}, while MAPK1 is also known to interact with MAPT \citep{Leugers2013}, SMAD3 \citep{Fang2012}, and BMP6 \citep{Zhang2018}.
Additional supporting evidence can be found in Appendix \ref{app:metabric}.

\vspace{-3mm}

\section{Conclusion}\label{sec:conclusion}

\vspace{-3mm}

In this paper, we introduced {\proposed}, an ensemble-based approach that tackles the newly proposed challenge of composite feature selection. 
Using synthetic and semi-synthetic data, we assess our ability to go beyond traditional feature selection and recover deeper underlying connections between variables. 
\proposed{} is not without limitations: as with other methods, points of difficulty arise when features are highly correlated, or if predictive composites contain overlapping features. Future work may overcome this by using correlated gates. Further, as with many traditional feature selection methods, there are no guarantees on false discovery rate. This could be tackled by first proposing candidate composite features, and then using the Group Knockoff procedure. Additionally, to discover groups, {\proposed} requires the introduction of additional hyperparameters which could be challenging to tune in practice. More broadly, as with standard feature selection, groups found under composite feature selection must be verified by domain experts (both features but additionally interactions). However, we believe the additional structure provided by composite feature selection could be of significant benefit to a wide variety of practitioners.

\section*{Acknowledgements}
We thank the anonymous reviewers for their comments and suggestions. We also thank Bogdan Cebere and Evgeny Saveliev for reviewing our public code. Fergus Imrie and Mihaela van der Schaar are supported by the National Science Foundation (NSF, grant number 1722516).
Mihaela van der Schaar is additionally supported by the Office of Naval Research (ONR). Alexander Norcliffe is supported by a GlaxoSmithKline grant.

\clearpage

\bibliography{main}

\begin{thebibliography}{100}

\bibitem{ALSARAKBI2010}
Wail Al~Sarakbi, Sara Reefy, Wen~G. Jiang, Terry Roberts, Robert~F. Newbold,
  and Kefah Mokbel.
\newblock Evidence of a tumour suppressor function for {DLEC1} in human breast
  cancer.
\newblock {\em Anticancer Research}, 30(4):1079--1082, 2010.

\bibitem{Amaldi1998approximability}
Edoardo Amaldi and Viggo Kann.
\newblock On the approximability of minimizing nonzero variables or unsatisfied
  relations in linear systems.
\newblock {\em Theoretical Computer Science}, 209(1):237--260, 1998.

\bibitem{AMPUJA2016}
M.~Ampuja, E.L. Alarmo, P.~Owens, R.~Havunen, A.E. Gorska, H.L. Moses, and
  A.~Kallioniemi.
\newblock The impact of bone morphogenetic protein 4 ({BMP4}) on breast cancer
  metastasis in a mouse xenograft model.
\newblock {\em Cancer Letters}, 375(2):238--244, 2016.

\bibitem{Arcuri2021}
Sharon Arcuri, Georgia Pennarossa, Fulvio Gandolfi, and Tiziana A.~L. Brevini.
\newblock Generation of trophoblast-like cells from hypomethylated porcine
  adult dermal fibroblasts.
\newblock {\em Frontiers in Veterinary Science}, 8, 2021.

\bibitem{Bakele2014}
Martina Bakele, Amelie~S. Lotz-Havla, Anja Jakowetz, Melanie Carevic, Veronica
  Marcos, Ania~C. Muntau, and Dominik Gersting, Soeren W.and~Hartl.
\newblock An interactive network of elastase, secretases, and {PAR}-2 protein
  regulates {CXCR1} receptor surface expression on neutrophils.
\newblock {\em Journal of Biological Chemistry}, 289(30):20516--20525, 2014.

\bibitem{Balendra2018}
Rubika Balendra and Adrian~M. Isaacs.
\newblock C9orf72-mediated {ALS} and {FTD}: multiple pathways to disease.
\newblock {\em Nature Reviews Neurology}, 14(9):544--558, Sep 2018.

\bibitem{abid2019concrete}
Muhammed~Fatih Bal{\i}n, Abubakar Abid, and James Zou.
\newblock Concrete autoencoders: Differentiable feature selection and
  reconstruction.
\newblock In {\em International Conference on Machine Learning (ICML)}, 2019.

\bibitem{Balli2013}
Hatice~Ozer Balli and Bent~E. S{\o}rensen.
\newblock Interaction effects in econometrics.
\newblock {\em Empirical Economics}, 45(1):583--603, 2013.

\bibitem{barber2015controlling}
Rina~Foygel Barber and Emmanuel~J Cand{\`e}s.
\newblock Controlling the false discovery rate via knockoffs.
\newblock {\em The Annals of Statistics}, 43(5):2055--2085, 2015.

\bibitem{beck2009fast}
Amir Beck and Marc Teboulle.
\newblock A fast iterative shrinkage-thresholding algorithm for linear inverse
  problems.
\newblock {\em SIAM Journal on Imaging Sciences}, 2(1):183--202, 2009.

\bibitem{bellot2019conditional}
Alexis Bellot and Mihaela van~der Schaar.
\newblock Conditional independence testing using generative adversarial
  networks.
\newblock In {\em Advances in Neural Information Processing Systems (NeurIPS)},
  2019.

\bibitem{bogdan2013statistical}
Malgorzata Bogdan, Ewout van~den Berg, Weijie Su, and Emmanuel Candes.
\newblock Statistical estimation and testing via the sorted {L1} norm.
\newblock {\em arXiv preprint arXiv:1310.1969}, 2013.

\bibitem{Brown2009}
Terry Brown.
\newblock {Silica exposure, smoking, silicosis and lung cancer—complex
  interactions}.
\newblock {\em Occupational Medicine}, 59(2):89--95, 03 2009.

\bibitem{brzyski2019group}
Damian Brzyski, Alexej Gossmann, Weijie Su, and Ma{\l}gorzata Bogdan.
\newblock Group {SLOPE} - adaptive selection of groups of predictors.
\newblock {\em Journal of the American Statistical Association},
  114(525):419--433, 2019.

\bibitem{Byerly2016}
Jessica Byerly, Gwyneth Halstead-Nussloch, Koichi Ito, Igor Katsyv, and
  Hanna~Y. Irie.
\newblock {PRKCQ} promotes oncogenic growth and anoikis resistance of a subset
  of triple-negative breast cancer cells.
\newblock {\em Breast Cancer Research}, 18(1):95, 2016.

\bibitem{Byerly2020}
Jessica~H. Byerly, Elisa~R. Port, and Hanna~Y. Irie.
\newblock {PRKCQ} inhibition enhances chemosensitivity of triple-negative
  breast cancer by regulating {B}im.
\newblock {\em Breast Cancer Research}, 22(1):72, 2020.

\bibitem{candes2018panning}
Emmanuel Candes, Yingying Fan, Lucas Janson, and Jinchi Lv.
\newblock Panning for gold: ‘model-{X}’ knockoffs for high dimensional
  controlled variable selection.
\newblock {\em Journal of the Royal Statistical Society: Series B (Statistical
  Methodology)}, 80(3):551--577, 2018.

\bibitem{Chandrashekar2014survey}
Girish Chandrashekar and Ferat Sahin.
\newblock A survey on feature selection methods.
\newblock {\em Computers \& Electrical Engineering}, 40(1):16--28, 2014.

\bibitem{Chen2022}
Bijun Chen, Ruoshui Li, Silvia~C. Hernandez, Anis Hanna, Kai Su, Arti~V.
  Shinde, and Nikolaos~G. Frangogiannis.
\newblock Differential effects of smad2 and smad3 in regulation of macrophage
  phenotype and function in the infarcted myocardium.
\newblock {\em Journal of Molecular and Cellular Cardiology}, 171:1--15, 2022.

\bibitem{Cheng2022}
Tianyi Cheng, Peiying Chen, Jingyi Chen, Yingtong Deng, and Chen Huang.
\newblock Landscape analysis of matrix metalloproteinases unveils key
  prognostic markers for patients with breast cancer.
\newblock {\em Frontiers in Genetics}, 12, 2022.

\bibitem{Curtis2012}
Christina Curtis, Sohrab~P. Shah, Suet-Feung Chin, Gulisa Turashvili, Oscar~M.
  Rueda, Mark~J. Dunning, Doug Speed, Andy~G. Lynch, Shamith Samarajiwa, Yinyin
  Yuan, Stefan Gr{\"a}f, Gavin Ha, Gholamreza Haffari, Ali Bashashati, Roslin
  Russell, Steven McKinney, METABRIC Group, Anita Langer{\o}d, Andrew Green,
  Elena Provenzano, Gordon Wishart, Sarah Pinder, Peter Watson, Florian
  Markowetz, Leigh Murphy, Ian Ellis, Arnie Purushotham, Anne-Lise
  B{\o}rresen-Dale, James~D. Brenton, Simon Tavar{\'e}, Carlos Caldas, and
  Samuel Aparicio.
\newblock The genomic and transcriptomic architecture of 2,000 breast tumours
  reveals novel subgroups.
\newblock {\em Nature}, 486(7403):346--352, 2012.

\bibitem{Dai:13}
Kun Dai, Hong-Yi Yu, and Qing Li.
\newblock A semisupervised feature selection with support vector machine.
\newblock {\em Journal of Applied Mathematics}, 64:141--158, 2013.

\bibitem{dai2016knockoff}
Ran Dai and Rina Barber.
\newblock The knockoff filter for {FDR} control in group-sparse and multitask
  regression.
\newblock In {\em International Conference on Machine Learning (ICML)}, 2016.

\bibitem{Mei2007}
Mei Dong, Tam How, Kellye~C. Kirkbride, Kelly~J. Gordon, Jason~D. Lee, Nadine
  Hempel, Patrick Kelly, Benjamin~J. Moeller, Jeffrey~R. Marks, and Gerard~C.
  Blobe.
\newblock The type {III} {TGF}-$\beta$ receptor suppresses breast cancer
  progression.
\newblock {\em The Journal of Clinical Investigation}, 117(1):206--217, 2007.

\bibitem{doran2014permutation}
Gary Doran, Krikamol Muandet, Kun Zhang, and Bernhard Sch{\"o}lkopf.
\newblock A permutation-based kernel conditional independence test.
\newblock In {\em Uncertainty in Artificial Intelligence (UAI)}, pages
  132--141, 2014.

\bibitem{Fang2012}
Wei~Bin Fang, Iman Jokar, An~Zou, Diana Lambert, Prasanthi Dendukuri, and Nikki
  Cheng.
\newblock {CCL2}/{CCR2} chemokine signaling coordinates survival and motility
  of breast cancer cells through smad3 protein- and p42/44 mitogen-activated
  protein kinase ({MAPK})-dependent mechanisms.
\newblock {\em Journal of Biological Chemistry}, 287(43):36593--36608, 2012.

\bibitem{Fessing2010}
Michael~Y. Fessing, Ruzanna Atoyan, Ben Shander, Andrei~N. Mardaryev, Vladimir
  V.~Botchkarev Jr., Krzysztof Poterlowicz, Yonghong Peng, Tatiana Efimova, and
  Vladimir~A. Botchkarev.
\newblock {BMP} signaling induces cell-type-specific changes in gene expression
  programs of human keratinocytes and fibroblasts.
\newblock {\em Journal of Investigative Dermatology}, 130(2):398--404, 2010.

\bibitem{george1997approaches}
Edward~I. George and Robert~E. McCulloch.
\newblock Approaches for bayesian variable selection.
\newblock {\em Statistica Sinica}, pages 339--373, 1997.

\bibitem{Ginestier2010}
Christophe Ginestier, Suling Liu, Mark~E. Diebel, Hasan Korkaya, Ming Luo,
  Marty Brown, Julien Wicinski, Olivier Cabaud, Emmanuelle Charafe-Jauffret,
  Daniel Birnbaum, Jun-Lin Guan, Gabriela Dontu, and Max~S. Wicha.
\newblock {CXCR1} blockade selectively targets human breast cancer stem cells
  in vitro and in xenografts.
\newblock {\em The Journal of Clinical Investigation}, 120(2):485--497, 2010.

\bibitem{GomezBergna2021}
Santiago~M. G{\'o}mez~Bergna, Abril Marchesini, Leslie C.~Amor{\'o}s Morales,
  Paula~N. Arr{\'\i}as, Hern{\'a}n~G. Farina, V{\'\i}ctor Romanowski,
  M.~Florencia Gottardo, and Matias~L. Pidre.
\newblock Exploring the metastatic role of the inhibitor of apoptosis {BIRC6}
  in breast cancer.
\newblock {\em bioRxiv}, 2021.

\bibitem{Graham2020}
Daniel~B. Graham and Ramnik~J. Xavier.
\newblock Pathway paradigms revealed from the genetics of inflammatory bowel
  disease.
\newblock {\em Nature}, 578(7796):527--539, Feb 2020.

\bibitem{Guyon2003introduction}
Isabelle Guyon and Andr{\'{e}} Elisseeff.
\newblock An introduction to variable and feature selection.
\newblock {\em Journal of Machine Learning Research}, 3(1):1157--1182, 2003.

\bibitem{Han2018}
Kai Han, Yunhe Wang, Chao Zhang, Chao Li, and Chao Xu.
\newblock Autoencoder inspired unsupervised feature selection.
\newblock In {\em IEEE International Conference on Acoustics, Speech and Signal
  Processing (ICASSP)}, 2018.

\bibitem{He2005laplacian}
Xiaofei He, Deng Cai, and Partha Niyogi.
\newblock Laplacian score for feature selection.
\newblock In {\em Advances in Neural Information Processing Systems (NeurIPS)},
  2005.

\bibitem{hernandez2013generalized}
Daniel Hern{\'a}ndez-Lobato, Jos{\'e}~Miguel Hern{\'a}ndez-Lobato, and Pierre
  Dupont.
\newblock Generalized spike-and-slab priors for bayesian group feature
  selection using expectation propagation.
\newblock {\em Journal of Machine Learning Research}, 14(7), 2013.

\bibitem{Irwin2012zinc}
John~J. Irwin, Teague Sterling, Michael~M. Mysinger, Erin~S. Bolstad, and
  Ryan~G. Coleman.
\newblock {ZINC}: {A} free tool to discover chemistry for biology.
\newblock {\em Journal of Chemical Information and Modeling}, 52(7):1757--1768,
  2012.

\bibitem{Jaccard1912}
Paul Jaccard.
\newblock The distribution of the flora in the alpine zone.
\newblock {\em The New Phytologist}, 11(2):37--50, 1912.

\bibitem{Jang17categorical}
Eric Jang, Shixiang Gu, and Ben Poole.
\newblock Categorical reparameterization with {G}umbel-{S}oftmax.
\newblock In {\em International Conference on Learning Representations (ICLR)},
  2017.

\bibitem{jordon2018knockoffgan}
James Jordon, Jinsung Yoon, and Mihaela van~der Schaar.
\newblock {KnockoffGAN}: Generating knockoffs for feature selection using
  generative adversarial networks.
\newblock In {\em International Conference on Learning Representations (ICLR)},
  2018.

\bibitem{Paivi2018}
Päivi Järvensivu, Taija Heinosalo, Janne Hakkarainen, Pauliina Kronqvist,
  Niina Saarinen, and Matti Poutanen.
\newblock {HSD17B1} expression induces inflammation-aided rupture of mammary
  gland myoepithelium.
\newblock {\em Endocrine-Related Cancer}, 25(4):393 -- 406, 2018.

\bibitem{Kato2018}
Tadashi Kato, Atsushi Yamada, Mikiko Ikehata, Yuko Yoshida, Kiyohito Sasa,
  Naoko Morimura, Akiko Sakashita, Takehiko Iijima, Daichi Chikazu, Hiroaki
  Ogata, and Ryutaro Kamijo.
\newblock {FGF}-2 suppresses expression of nephronectin via {JNK} and {PI3K}
  pathways.
\newblock {\em FEBS Open Bio}, 8(5):836--842, 2018.

\bibitem{Kelly2009multiple}
Matthew Kelly and Christopher Semsarian.
\newblock Multiple mutations in genetic cardiovascular disease.
\newblock {\em Circulation: Cardiovascular Genetics}, 2(2):182--190, 2009.

\bibitem{Kim2015}
Pora Kim, Feixiong Cheng, Junfei Zhao, and Zhongming Zhao.
\newblock {ccmGDB}: a database for cancer cell metabolism genes.
\newblock {\em Nucleic Acids Research}, 44(D1):D959--D968, 2015.

\bibitem{kingma2014adam}
Diederik~P. Kingma and Jimmy Ba.
\newblock Adam: A method for stochastic optimization.
\newblock {\em arXiv preprint arXiv:1412.6980}, 2014.

\bibitem{kira1992practical}
Kenji Kira and Larry~A. Rendell.
\newblock A practical approach to feature selection.
\newblock In {\em Machine Learning Proceedings}, pages 249--256. 1992.

\bibitem{Knijnenburg2016logic}
Theo~A. Knijnenburg, Gunnar~W. Klau, Francesco Iorio, Mathew~J. Garnett, Ultan
  McDermott, Ilya Shmulevich, and Lodewyk F.~A. Wessels.
\newblock Logic models to predict continuous outputs based on binary inputs
  with an application to personalized cancer therapy.
\newblock {\em Scientific Reports}, 6(1):36812, 2016.

\bibitem{Kohavi1997wrapper}
Ron Kohavi and George~H. John.
\newblock Wrappers for feature subset selection.
\newblock {\em Artificial Intelligence}, 97(1):273--324, 1997.

\bibitem{lecun1998gradient}
Yann LeCun, L{\'e}on Bottou, Yoshua Bengio, and Patrick Haffner.
\newblock Gradient-based learning applied to document recognition.
\newblock {\em Proceedings of the IEEE}, 86(11):2278--2324, 1998.

\bibitem{lee2021self}
Changhee Lee, Fergus Imrie, and Mihaela van~der Schaar.
\newblock Self-supervision enhanced feature selection with correlated gates.
\newblock In {\em International Conference on Learning Representations (ICLR)},
  2022.

\bibitem{Lemhadri2021}
Ismael Lemhadri, Feng Ruan, and Rob Tibshirani.
\newblock {LassoNet}: Neural networks with feature sparsity.
\newblock In {\em International Conference on Artificial Intelligence and
  Statistics (AISTATS)}, 2021.

\bibitem{Leugers2013}
Chad Leugers, Ju~Yong Koh, Willis Hong, and Gloria Lee.
\newblock Tau in {MAPK} activation.
\newblock {\em Frontiers in Neurology}, 4, 2013.

\bibitem{Li2014}
Xinghua Li, Weijiang Liang, Junling Liu, Chuyong Lin, Shu Wu, Libing Song, and
  Zhongyu Yuan.
\newblock Transducin ($\beta$)-like 1 {X}-linked receptor 1 promotes
  proliferation and tumorigenicity in human breast cancer via activation of
  beta-catenin signaling.
\newblock {\em Breast Cancer Research}, 16(5):465, 2014.

\bibitem{Li:16_DeepFS}
Yifeng~Li Li, Chih-Yu Chen, and Wyeth~W. Wasserman.
\newblock Deep feature selection: theory and application to identify enhancers
  and promoters.
\newblock {\em Journal of Computational Biology}, 23(5):322--336, 2016.

\bibitem{Liang18bayesian}
Faming Liang, Qizhai Li, and Lei Zhou.
\newblock Bayesian neural networks for selection of drug sensitive genes.
\newblock {\em Journal of the American Statistical Association},
  113(523):955--972, 2018.

\bibitem{Lindenbaum2021}
Ofir Lindenbaum, Uri Shaham, Erez Peterfreund, Jonathan Svirsky, Nicolas Casey,
  and Yuval Kluger.
\newblock Differentiable unsupervised feature selection based on a gated
  laplacian.
\newblock In {\em Advances in Neural Information Processing Systems (NeurIPS)},
  2021.

\bibitem{Liu1996filter}
Huan Liu and Rudy Setiono.
\newblock A probabilistic approach to feature selection - a filter solution.
\newblock In {\em International Conference on Machine Learning (ICML)}, 1996.

\bibitem{liu2019deep}
Ying Liu and Cheng Zheng.
\newblock Deep latent variable models for generating knockoffs.
\newblock {\em Stat}, 8(1):e260, 2019.

\bibitem{Lu2019}
Huanyu Lu, Yue Guo, Gaurav Gupta, and Xingsong Tian.
\newblock Mitogen-activated protein kinase ({MAPK}): New insights in breast
  cancer.
\newblock {\em Journal of Environmental Pathology, Toxicology and Oncology},
  38(1):51--59, 2019.

\bibitem{Maddison17concrete}
Chris~J. Maddison, Andriy Mnih, and Yee~Whye Teh.
\newblock The {C}oncrete distribution: A continuous relaxation of discrete
  random variables.
\newblock In {\em International Conference on Learning Representations (ICLR)},
  2017.

\bibitem{Mani2008}
Ramamurthy Mani, Robert~P. St.Onge, John~L. Hartman, Guri Giaever, and
  Frederick~P. Roth.
\newblock Defining genetic interaction.
\newblock {\em Proceedings of the National Academy of Sciences},
  105(9):3461--3466, 2008.

\bibitem{Manna2019}
Pulak~R. Manna, Ahsen~U. Ahmed, Shengping Yang, Madhusudhanan Narasimhan,
  Joëlle Cohen-Tannoudji, Andrzej~T. Slominski, and Kevin Pruitt.
\newblock Genomic profiling of the steroidogenic acute regulatory protein in
  breast cancer: In silico assessments and a mechanistic perspective.
\newblock {\em Cancers}, 11(5), 2019.

\bibitem{McCloskey2019attribution}
Kevin McCloskey, Ankur Taly, Federico Monti, Michael~P. Brenner, and Lucy~J.
  Colwell.
\newblock Using attribution to decode binding mechanism in neural network
  models for chemistry.
\newblock {\em Proceedings of the National Academy of Sciences},
  116(24):11624--11629, 2019.

\bibitem{Meira2001}
Lisiane~B. Meira, Antonio~M.C. Reis, David~L. Cheo, Dorit Nahari, Dennis~K.
  Burns, and Errol~C. Friedberg.
\newblock Cancer predisposition in mutant mice defective in multiple genetic
  pathways: uncovering important genetic interactions.
\newblock {\em Mutation Research/Fundamental and Molecular Mechanisms of
  Mutagenesis}, 477(1):51--58, 2001.

\bibitem{Merino2022}
Jordi Merino, Marta Guasch-Ferré, Jun Li, Wonil Chung, Yang Hu, Baoshan Ma,
  Yanping Li, Jae~H. Kang, Peter Kraft, Liming Liang, Qi~Sun, Paul~W. Franks,
  JoAnn~E. Manson, Walter~C. Willet, Jose~C. Florez, and Frank~B. Hu.
\newblock Polygenic scores, diet quality, and type 2 diabetes risk: An
  observational study among 35,759 adults from 3 {US} cohorts.
\newblock {\em PLOS Medicine}, 19(4):1--20, 04 2022.

\bibitem{Papadimitriou2019predicting}
Sofia Papadimitriou, Andrea Gazzo, Nassim Versbraegen, Charlotte Nachtegael,
  Jan Aerts, Yves Moreau, Sonia Van~Dooren, Ann Now{\'e}, Guillaume Smits, and
  Tom Lenaerts.
\newblock Predicting disease-causing variant combinations.
\newblock {\em Proceedings of the National Academy of Sciences},
  116(24):11878--11887, 2019.

\bibitem{Park2016}
Ui-Hyun Park, Mi~Ran Kang, Eun-Joo Kim, Young-Soo Kwon, Wooyoung Hur, Seung~Kew
  Yoon, Byoung-Joon Song, Jin~Hwan Park, Jin-Taek Hwang, Ji-Cheon Jeong, and
  Soo-Jong Um.
\newblock {ASXL2} promotes proliferation of breast cancer cells by linking
  {ER$\alpha$} to histone methylation.
\newblock {\em Oncogene}, 35(28):3742--3752, 2016.

\bibitem{Peltonen2013}
Hanna~M. Peltonen, Annakaisa Haapasalo, Mikko Hiltunen, Vesa Kataja, Veli-Matti
  Kosma, and Arto Mannermaa.
\newblock {$\Gamma$}-secretase components as predictors of breast cancer
  outcome.
\newblock {\em PLOS ONE}, 8(11), 2013.

\bibitem{Pereira2016}
Bernard Pereira, Suet-Feung Chin, Oscar~M. Rueda, Hans-Kristian~Moen Vollan,
  Elena Provenzano, Helen~A. Bardwell, Michelle Pugh, Linda Jones, Roslin
  Russell, Stephen-John Sammut, Dana W.~Y. Tsui, Bin Liu, Sarah-Jane Dawson,
  Jean Abraham, Helen Northen, John~F. Peden, Abhik Mukherjee, Gulisa
  Turashvili, Andrew~R. Green, Steve McKinney, Arusha Oloumi, Sohrab Shah,
  Nitzan Rosenfeld, Leigh Murphy, David~R. Bentley, Ian~O. Ellis, Arnie
  Purushotham, Sarah~E. Pinder, Anne-Lise B{\o}rresen-Dale, Helena~M. Earl,
  Paul~D. Pharoah, Mark~T. Ross, Samuel Aparicio, and Carlos Caldas.
\newblock The somatic mutation profiles of 2,433 breast cancers refine their
  genomic and transcriptomic landscapes.
\newblock {\em Nature Communications}, 7(1):11479, 2016.

\bibitem{Phillips2008}
Patrick~C. Phillips.
\newblock Epistasis --- the essential role of gene interactions in the
  structure and evolution of genetic systems.
\newblock {\em Nature Reviews Genetics}, 9(11):855--867, 2008.

\bibitem{Pickup2015}
Michael~W. Pickup, Laura~D. Hover, Eleanor~R. Polikowsky, Anna Chytil,
  Agnieszka~E. Gorska, Sergey~V. Novitskiy, Harold~L. Moses, and Philip Owens.
\newblock {BMPR2} loss in fibroblasts promotes mammary carcinoma metastasis via
  increased inflammation.
\newblock {\em Molecular Oncology}, 9(1):179--191, 2015.

\bibitem{Piskor2020}
Barbara~Maria Piskór, Andrzej Przylipiak, Emilia Dąbrowska, Iwona
  Sidorkiewicz, Marek Niczyporuk, Maciej Szmitkowski, and Sławomir Ławicki.
\newblock Plasma level of {MMP}-10 may be a prognostic marker in early stages
  of breast cancer.
\newblock {\em Journal of Clinical Medicine}, 9(12), 2020.

\bibitem{Rapaport2008classification}
Franck Rapaport, Emmanuel Barillot, and Jean-Philippe Vert.
\newblock Classification of array{CGH} data using fused {SVM}.
\newblock {\em Bioinformatics}, 24(13):i375--i382, 2008.

\bibitem{romano2020deep}
Yaniv Romano, Matteo Sesia, and Emmanuel Cand{\`e}s.
\newblock Deep knockoffs.
\newblock {\em Journal of the American Statistical Association},
  115(532):1861--1872, 2020.

\bibitem{runge2018conditional}
Jakob Runge.
\newblock Conditional independence testing based on a nearest-neighbor
  estimator of conditional mutual information.
\newblock In {\em International Conference on Artificial Intelligence and
  Statistics (AISTATS)}, pages 938--947, 2018.

\bibitem{Sanchez-Lengeling2020evaluating}
Benjamin Sanchez-Lengeling, Jennifer Wei, Brian Lee, Emily Reif, Peter Wang,
  Wesley Qian, Kevin McCloskey, Lucy Colwell, and Alexander Wiltschko.
\newblock Evaluating attribution for graph neural networks.
\newblock In {\em Advances in Neural Information Processing Systems (NeurIPS)},
  2020.

\bibitem{santosa1986linear}
Fadil Santosa and William~W. Symes.
\newblock Linear inversion of band-limited reflection seismograms.
\newblock {\em SIAM Journal on Scientific and Statistical Computing},
  7(4):1307--1330, 1986.

\bibitem{sen2017model}
Rajat Sen, Ananda~Theertha Suresh, Karthikeyan Shanmugam, Alexandros~G Dimakis,
  and Sanjay Shakkottai.
\newblock Model-powered conditional independence test.
\newblock In {\em Advances in neural information processing systems (NeurIPS)},
  2017.

\bibitem{Sheikhpour:17}
Razieh Sheikhpour, Mehdi~Agha Sarram, Sajjad Gharaghani, and Mohammad Ali~Zare
  Chahooki.
\newblock A survey on semi-supervised feature selection methods.
\newblock {\em Pattern Recognition}, 64:141--158, 2017.

\bibitem{singha2019increased}
Prajjal~K. Singha, Srilakshmi Pandeswara, Hui Geng, Rongpei Lan, Manjeri~A.
  Venkatachalam, Albert Dobi, Shiv Srivastava, and Pothana Saikumar.
\newblock Increased smad3 and reduced smad2 levels mediate the functional
  switch of {TGF-$\beta$} from growth suppressor to growth and metastasis
  promoter through {TMEPAI/PMEPA1} in triple negative breast cancer.
\newblock {\em Genes \& cancer}, 10(5-6):134, 2019.

\bibitem{sudarshan2020deep}
Mukund Sudarshan, Wesley Tansey, and Rajesh Ranganath.
\newblock Deep direct likelihood knockoffs.
\newblock In {\em Advances in Neural Information Processing Systems (NeurIPS)},
  2020.

\bibitem{Takahashi2008}
Mina Takahashi, Fumio Otsuka, Tomoko Miyoshi, Hiroyuki Otani, Junko Goto,
  Misuzu Yamashita, Toshio Ogura, Hirofumi Makino, and Hiroyoshi Doihara.
\newblock Bone morphogenetic protein 6 ({BMP6}) and {BMP7} inhibit
  estrogen-induced proliferation of breast cancer cells by suppressing p38
  mitogen-activated protein kinase activation.
\newblock {\em Journal of Endocrinology}, 199(3):445 -- 455, 2008.

\bibitem{tang2014feature}
Jiliang Tang, Salem Alelyani, and Huan Liu.
\newblock Feature selection for classification: A review.
\newblock {\em Data classification: Algorithms and applications}, page~37,
  2014.

\bibitem{tibshirani1996regression}
Robert Tibshirani.
\newblock Regression shrinkage and selection via the lasso.
\newblock {\em Journal of the Royal Statistical Society: Series B
  (Methodological)}, 58(1):267--288, 1996.

\bibitem{Turner2010}
Nicholas Turner, Alex Pearson, Rachel Sharpe, Maryou Lambros, Felipe Geyer,
  Maria~A. Lopez-Garcia, Rachael Natrajan, Caterina Marchio, Elizabeth Iorns,
  Alan Mackay, Cheryl Gillett, Anita Grigoriadis, Andrew Tutt, Jorge~S.
  Reis-Filho, and Alan Ashworth.
\newblock {FGFR1} amplification drives endocrine therapy resistance and is a
  therapeutic target in breast cancer.
\newblock {\em Cancer Research}, 70(5):2085--2094, 2010.

\bibitem{Wang2019}
Dongfeng Wang, Jian Li, Fengling Cai, Zhi Xu, Li~Li, Huanfeng Zhu, Wei Liu,
  Qingyu Xu, Jian Cao, Jingfeng Sun, and Jinhai Tang.
\newblock Overexpression of {MAPT-AS1} is associated with better patient
  survival in breast cancer.
\newblock {\em Biochemistry and Cell Biology}, 97(2):158--164, 2019.

\bibitem{Wang2018}
Dongsheng Wang, Chenglong Zhao, Liangliang Gao, Yao Wang, Xin Gao, Liang Tang,
  Kun Zhang, Zhenxi Li, Jing Han, and Jianru Xiao.
\newblock {NPNT} promotes early-stage bone metastases in breast cancer by
  regulation of the osteogenic niche.
\newblock {\em Journal of Bone Oncology}, 13:91--96, 2018.

\bibitem{wei2015expression}
Chang-Yuan Wei, Qi-Xing Tan, Xiao Zhu, Qing-Hong Qin, Fei-Bai Zhu, Qin-Guo Mo,
  and Wei-Ping Yang.
\newblock Expression of {CDKN1A/p21} and {TGFBR2} in breast cancer and their
  prognostic significance.
\newblock {\em International Journal of Clinical and Experimental Pathology},
  8(11):14619, 2015.

\bibitem{Wendt2014}
Michael~K. Wendt, Molly~A. Taylor, Barbara~J. Schiemann, Khalid Sossey-Alaoui,
  and William~P. Schiemann.
\newblock Fibroblast growth factor receptor splice variants are stable markers
  of oncogenic transforming growth factor $\beta$1 signaling in metastatic
  breast cancers.
\newblock {\em Breast Cancer Research}, 16(2):R24, 2014.

\bibitem{Wilson2001}
Tim Wilson, Tim Holt, and Trisha Greenhalgh.
\newblock Complexity and clinical care.
\newblock {\em BMJ}, 323(7314):685--688, 2001.

\bibitem{Wu2021}
Xinxing Wu and Qiang Cheng.
\newblock Algorithmic stability and generalization of an unsupervised feature
  selection algorithm.
\newblock In {\em Advances in Neural Information Processing Systems (NeurIPS)},
  2021.

\bibitem{Yamada2020feature}
Yutaro Yamada, Ofir Lindenbaum, Sahand Negahban, and Yuval Kluger.
\newblock Feature selection using stochastic gates.
\newblock In {\em International Conference on Machine Learning (ICML)}, 2020.

\bibitem{yoon2018invase}
Jinsung Yoon, James Jordon, and Mihaela van~der Schaar.
\newblock Invase: Instance-wise variable selection using neural networks.
\newblock In {\em International Conference on Learning Representations (ICLR)},
  2019.

\bibitem{yuan2006model}
Ming Yuan and Yi~Lin.
\newblock Model selection and estimation in regression with grouped variables.
\newblock {\em Journal of the Royal Statistical Society: Series B (Statistical
  Methodology)}, 68(1):49--67, 2006.

\bibitem{zaheer2017deep}
Manzil Zaheer, Satwik Kottur, Siamak Ravanbakhsh, Barnabas Poczos, Ruslan
  Salakhutdinov, and Alexander Smola.
\newblock Deep sets.
\newblock In {\em Advances in Neural Information Processing Systems (NeurIPS)},
  2017.

\bibitem{zeng2014group}
Lingmin Zeng and Jun Xie.
\newblock Group variable selection via {SCAD-L}2.
\newblock {\em Statistics}, 48(1):49--66, 2014.

\bibitem{Zhang2018}
Xin-Yue Zhang, Hsun-Ming Chang, Elizabeth~L. Taylor, Rui-Zhi Liu, and Peter
  C.~K. Leung.
\newblock {BMP6} downregulates {GDNF} expression through {SMAD1}/5 and {ERK1}/2
  signaling pathways in human granulosa-lutein cells.
\newblock {\em Endocrinology}, 159(8):2926--2938, 2018.

\bibitem{Zhang2021}
Yong-ping Zhang, Wen-ting Na, Xiao-qiang Dai, Ruo-fei Li, Jian-xiong Wang, Ting
  Gao, Wei-bo Zhang, and Cheng Xiang.
\newblock Over-expression of {SRD5A3} and its prognostic significance in breast
  cancer.
\newblock {\em World Journal of Surgical Oncology}, 19(1):260, 2021.

\bibitem{Zhao:08}
Jidong Zhao, Ke~Lu, and Xiaofei He.
\newblock Locality sensitive semi-supervised feature selection.
\newblock {\em Neurocomputing}, 71:1842–--1849, 2008.

\bibitem{ZHONG2015}
Ting Zhong, Feifei Xu, Jinhui Xu, Liang Liu, and Yun Chen.
\newblock Aldo-keto reductase {1C3} ({AKR1C3}) is associated with the
  doxorubicin resistance in human breast cancer via {PTEN} loss.
\newblock {\em Biomedicine \& Pharmacotherapy}, 69:317--325, 2015.

\bibitem{zhou2010group}
Nengfeng Zhou and Ji~Zhu.
\newblock Group variable selection via a hierarchical lasso and its oracle
  property.
\newblock {\em arXiv preprint arXiv:1006.2871}, 2010.

\bibitem{zhu2021deep}
Guangyu Zhu and Tingting Zhao.
\newblock Deep-g{K}nock: Nonlinear group-feature selection with deep neural
  networks.
\newblock {\em Neural Networks}, 135:139--147, 2021.

\end{thebibliography}
\bibliographystyle{plain}

\clearpage

\section*{Checklist}

\begin{enumerate}

\item For all authors...
\begin{enumerate}
  \item Do the main claims made in the abstract and introduction accurately reflect the paper's contributions and scope?
    \answerYes{}
  \item Did you describe the limitations of your work?
    \answerYes{See Conclusion for details.}
  \item Did you discuss any potential negative societal impacts of your work?
    \answerYes{In the Conclusion, we caution that as with any feature selection method, discovered features must be verified or evaluated by domain experts. This verification or evaluation might be costly, and should the method perform poorly, could result in wasted resources. In addition, without additional oversight (primarily in dataset construction but also when validating features), features that contain bias could remain and be identified by feature selection algorithms.}
  \item Have you read the ethics review guidelines and ensured that your paper conforms to them?
    \answerYes{}
\end{enumerate}

\item If you are including theoretical results...
\begin{enumerate}
  \item Did you state the full set of assumptions of all theoretical results?
    \answerNA{}
        \item Did you include complete proofs of all theoretical results?
    \answerNA{}
\end{enumerate}

\item If you ran experiments...
\begin{enumerate}
  \item Did you include the code, data, and instructions needed to reproduce the main experimental results (either in the supplemental material or as a URL)?
    \answerYes{Code is available at either of the following GitHub repositories: \url{https://github.com/a-norcliffe/Composite-Feature-Selection}, \url{https://github.com/vanderschaarlab/Composite-Feature-Selection}.}
  \item Did you specify all the training details (e.g., data splits, hyperparameters, how they were chosen)?
    \answerYes{Hyperparameters for each experiment are provided in Table \ref{Table:hyperparams}. Architecture details are provided in Appendix \ref{app:architecture} and further experimental details are provided in Appendix \ref{app:experiments}.}
        \item Did you report error bars (e.g., with respect to the random seed after running experiments multiple times)?
    \answerYes{All experiments are repeated 10 times and results are reported along with standard deviations.}
        \item Did you include the total amount of compute and the type of resources used (e.g., type of GPUs, internal cluster, or cloud provider)?
    \answerYes{All experiments can be run easily on a commercially-available laptop. We provide further details of the compute resources used in Appendix \ref{app:experiments}.}
\end{enumerate}

\item If you are using existing assets (e.g., code, data, models) or curating/releasing new assets...
\begin{enumerate}
  \item If your work uses existing assets, did you cite the creators?
    \answerYes{We used several existing methods and datasets (see Experiments). All benchmark methods and datasets are clearly cited.}
  \item Did you mention the license of the assets?
    \answerYes{Licenses of assets (benchmark methods and datasets) is provided in Appendix \ref{app:experiments} and \ref{app:chem}.}
  \item Did you include any new assets either in the supplemental material or as a URL?
    \answerYes{Code is available at either of the following GitHub repositories: \url{https://github.com/a-norcliffe/Composite-Feature-Selection}, \url{https://github.com/vanderschaarlab/Composite-Feature-Selection}.}
  \item Did you discuss whether and how consent was obtained from people whose data you're using/curating?
    \answerYes{We only use publicly available, anonymized datasets. See Broader Impact - Datasets.}
  \item Did you discuss whether the data you are using/curating contains personally identifiable information or offensive content?
    \answerYes{We only use publicly available, anonymized datasets. See Broader Impact - Datasets.}
\end{enumerate}

\item If you used crowdsourcing or conducted research with human subjects...
\begin{enumerate}
  \item Did you include the full text of instructions given to participants and screenshots, if applicable?
    \answerNA{}
  \item Did you describe any potential participant risks, with links to Institutional Review Board (IRB) approvals, if applicable?
    \answerNA{}
  \item Did you include the estimated hourly wage paid to participants and the total amount spent on participant compensation?
    \answerNA{}
\end{enumerate}

\end{enumerate}

\clearpage

\appendix

\section*{Broader Impact}

\paragraph{Applications.} Our work is focused on imposing additional structure on the traditional feature selection procedure. Feature selection and group feature selection help to improve a model's interpretability, from this we do not foresee negative impacts. One potentially significant issue is if biased data is used, feature selection and group feature selection may select features that are based on or serve to reinforce societal bias. Therefore, as mentioned in \ref{sec:conclusion} (Conclusion), discovered features must be verified by domain experts to avoid this. 

On the other hand, methods for understanding the importance of and interplay between features (such as our work) have potential for good. For example, if one were to discover the interactions between genes that leads to a particular disease this could greatly aid both screening and drug design. Other examples may be understanding complex interactions in economics, helping to understand how to tackle inequality; or interactions in climate science, helping to further our understanding of climate change. 
With respect to both positive and negative impacts, our work focuses more on introducing the problem of group feature selection, thus we do not see significant negative impact as a direct result of this paper. 

\paragraph{Datasets.} Three of our datasets do not contain sensitive data. These are the synthetic datasets, semi-synthetic chemistry datasets, and MNIST. We also use METABRIC, a dataset consisting of patients with breast cancer, containing genetic and medical information. Whilst this does contain sensitive information, this is a public dataset with anonymous data.

\section{Ablation Study}
\label{app:ablation}

\begin{figure}[h]
    \centering
    \includegraphics[width=\textwidth]{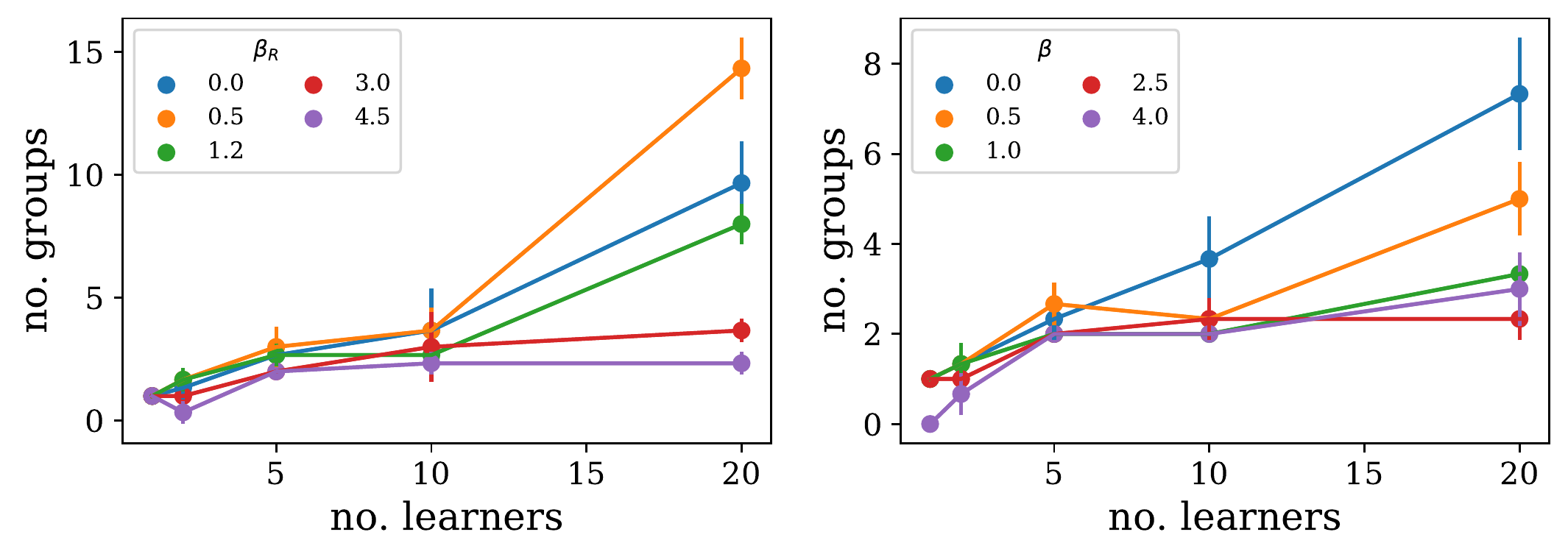}
    \caption{Ablation investigating how the number of learners in the ensemble and $\beta_R$ \& $\beta$ hyperparameters affect the number of groups learnt by \proposed.}
    \label{fig:ngroups}
\end{figure}

To explore the behavior of \proposed{} further, we conducted a sensitivity analysis of several hyperparameters. In particular, we investigated how many groups are used as the numbers of learners, $\beta_R$ and $\beta$ (scaling of $\langle\pi_i \rangle^2$) vary. 
As discussed in Section \ref{sec:exp}, the true number of composite features or groups is likely not known \textit{a priori}, and thus models for composite feature selection must not depend too heavily on this.
We examined how many learners are used (i.e. how many learners discover non-empty groups) on the Syn2 dataset. When ablating over $\beta_R$, we trained for 35 epochs, with $\beta$ fixed at 1.0. When ablating over $\beta$ we trained for 35 epochs keeping $\beta_R$ fixed at 1.2. Each experiment was repeated 3 times. The results are shown in Figure \ref{fig:ngroups}. For larger values of $\beta_R$ and $\beta$, the number of groups remains low, even when the number of learners far exceeds the true number of groups. As a consequence, if more learners are employed, larger values of $\beta_R$ and $\beta$ will likely be necessary.

\section{Extended Related Work}
\label{app:ext_related_work}

Feature selection has an extensive history and proposed solutions can be generally categorized into three types based on how they interact with the learning method: wrapper \citep{Kohavi1997wrapper}, filter \citep{Liu1996filter,kira1992practical}, or embedded methods \citep{santosa1986linear}. For more detailed discussion on traditional approaches to feature selection, we refer the reader to the following review articles \cite{tang2014feature,Chandrashekar2014survey}. In this section, we focus on advances in standard feature selection, which, while not the aim of our paper, provides useful background. 

Recently, a number of deep learning embedded feature selection methods have been developed which exploit end-to-end learning to jointly learn a prediction network that only uses a subset of the available features. 
For example, the popular LASSO penalization \cite{tibshirani1996regression} has been extended to neural networks in LassoNet \cite{Lemhadri2021}.  
One key challenge in deep learning approaches to feature selection is that selecting features is inherently a non-differentiable process.
Methods have explored different approximation schemes, such as Lasso or elastic net penalization (DeepFS) \citep{Li:16_DeepFS}, MCMC sampling (BNNsel) \citep{Liang18bayesian}, continuous relaxation using independent Gaussian random variables (STG) \citep{Yamada2020feature}, and reinforcement learning \citep{yoon2018invase}

In the unsupervised domain, differentiable unsupervised feature selection (DUFS) \citep{Lindenbaum2021} introduces a trainable Bernoulli gating mechanism into the Laplacian score \citep{He2005laplacian}, while \cite{Wu2021} extends autoencoder-based feature selection \cite{Han2018} with algorithmic stability.
Concrete Autoencoders (CAE) were proposed in \cite{abid2019concrete}, where autoencoders were employed to identify a pre-specified number of features that are sufficient for reconstructing the data. They employ a concrete selector layer based on Concrete random variables \cite{Maddison17concrete,Jang17categorical}. By selecting stochastic linear combinations of the input variables during training, the layer converges to a discrete set of features, with the number of feature defined by the user. While CAEs typically use a reconstruction objective (i.e. unsupervised), they can also be trained in a supervised manner.

Additionally, several semi-supervised feature selection methods have been proposed \citep{Sheikhpour:17} as extensions of traditional methods (e.g. \citep{Zhao:08,Dai:13}), as well as self-supervised methods \citep{lee2021self}. 

Finally, while not the focus of this paper, we briefly mention an alternate strand of literature related to feature selection, that of hypothesis testing and, in particular, conditional independence testing \cite{doran2014permutation,sen2017model,runge2018conditional,bellot2019conditional}.

\section{Architecture Details}
\label{app:architecture}

{\proposed} consists of 4 different network components: (i) group selection probabilities $\bm{\pi}_i$, (ii) group encoders $f_{\theta_i}$, (iii) group predictors $h_{\phi_i}$, and (iv) aggregate or ensemble predictor $h_{\phi}$. We use a fully-connected network as the baseline architecture for network components (ii)-(iv).
Note that compared to traditional feature selection, the computational complexity/number of parameters scales linearly with the maximum permitted number of groups (in the same way that an ensemble scales with the number of members). Typically the number of groups will be relatively small, which should alleviate any practical concerns. 
The specific architectural details of {\proposed} are provided below: 

\begin{itemize}[leftmargin=5.0mm]
\setlength{\itemsep}{0pt}
    \item Stochastic Gating: For the relaxed Bernoulli sampling we use a temperature of 0.1 and for evaluation we use a threshold $\lambda=0.7$ to determine whether a feature is relevant for the group.
    \item Group Encoders $f_{\theta_i}$: Multi-layer perceptron with ReLU activations, and two hidden layers of a given hidden width.
    \item Group Predictors $h_{\phi_i}$: Single linear layer, given by a separate weight matrix and bias for each group.
    \item Ensemble Predictor $h_{\phi}$: The latent representation of each group encoder is passed through a separate linear layer, given by a weight matrix and bias term. These are then summed element-wise, yielding a permutation invariant aggregation. For simplicity we do not apply any further transformations in our experiments.
\end{itemize}

Further details regarding hyperparameters in each of the experiments can be found in Table \ref{Table:hyperparams}.

\section{Experimental Details}
\label{app:experiments}

\paragraph{Benchmark implementations.} LASSO was implemented in PyTorch. Group LASSO was implemented using the \textit{group-lasso} public library \footnote{The documentation can be found at: \url{https://group-lasso.readthedocs.io/en/latest/}}, which is provided under the MIT License. This requires hyperparameter $\beta$, determining the trade-off between sparse features and predictive power. Group LASSO is trained with the fast iterative soft thresholding algorithm (FISTA) \cite{beck2009fast}. In order to run Group LASSO, we first form all possible combinations of groups of features, enumerating all groups of size 1 and 2. This allows us to treat the groups independently and run Group LASSO. This scales poorly requiring us to keep groups limited to a maximum size of 2. Additionally we eliminate the majority of the noise variables and use only 60 features in the synthetic experiments, giving Group LASSO a considerably easier task to tackle. Despite this, Group LASSO is still outperformed by \proposed. We use the publicly available implementations for both STG\footnote{\url{https://github.com/runopti/stg}} and CAE\footnote{\url{https://github.com/mfbalin/Concrete-Autoencoders}}. Both STG and CAE are provided under MIT Licenses.

\paragraph{Hyperparameters.} 
The hyperparameters used by each model in each experiment are given in Table \ref{Table:hyperparams}. In Table \ref{Table:hyperparams}, Reg refers to the regularisation parameter as follows. For {\proposed}, Reg refers to $\beta$. For both LASSO and Group LASSO, Reg refers to the scale factor that the $L1$ norm of the weights is multiplied by in the loss (often referred to as $\lambda$). For STG, Reg refers to $\lambda$, the scale factor for the $L1$ norm. For CAE, Reg refers to the number of features to use. Additionally for {\proposed}, $\beta$ and $\beta_{R}$ in this table are multiplied by the square root of the number of features for a given experiment. This allows the importance of having small groups and minimal overlap to increase with the dimensionality of the problem, but not too quickly. As a result, we observe that the hyperparameters are relatively stable across our experiments.
We fixed the number of possible groups for {\proposed} as five for all experiments. This is \textit{more} than the true number of groups, which means that {\proposed} must not only learn to select (composite) features, but also not to discover redundant features.
STG uses two hidden layers. CAE uses two hidden layers in the decoder.

\begin{table*}[h]
	\vskip -0.05in
	\caption{Hyperparameters for each experiment.} \label{Table:hyperparams}
	\vskip 0.15in
	\begin{center}
    \begin{small}
    \begin{sc}
		\begin{tabular}[b]{c c c c c c c c c}
			\toprule
			\multirow{2}{*}{\textbf{Task}} &
			\multirow{2}{*}{\textbf{Model}} & \multirow{2}{*}{Reg} & \multirow{2}{*}{$\beta_E$} & \multirow{2}{*}{$\beta_R$} & \textbf{Batch} & \textbf{Hidden} & \textbf{Learning} & \textbf{LR}\\
			& & & & & \textbf{Size} & \textbf{Width} & \textbf{Rate} & \textbf{Decay} \\
			\midrule
			\multirow{6}{*}{Syn1}
            & Compfs(5) & 4.5 & 1.0 & 1.2 & 50 & 20 & 0.003 & 0.99 \\
            & Compfs(1) & 0.35 & 1.0 & - & 100 & 30 & 0.003 & 0.99\\
            & LASSO & 0.4 & - & - & 50 & - & 0.003 & 0.99\\
            & Group LASSO & 0.1 & - & - & Full & - & - & - \\
            & STG & 0.1 & - & - & 250 & 20 & 0.001 & 1.00 \\
            & CAE & 2 & - & - & 250 & 200 & 0.001 & 1.00 \\
            \midrule
			\multirow{6}{*}{Syn2}
            & Compfs(5) & 4.5 & 1.0 & 1.2 & 50 & 20 & 0.003 & 0.99\\
            & Compfs(1) & 0.35 & 1.0 & - & 100 & 30 & 0.003 & 0.99\\
            & LASSO & 0.4 & - & - & 50 & - & 0.003 & 0.99\\
            & Group LASSO & 0.04 & - & - & Full & - & - & - \\
            & STG & 0.1 & - & - & 250 & 20 & 0.001 & 1.00 \\
            & CAE & 4 & - & - & 250 & 200 & 0.001 & 1.00 \\
            \midrule
			\multirow{6}{*}{Syn3}
            & Compfs(5) & 4.5 & 1.0 & 1.2 & 50 & 20 & 0.003 & 0.99\\
            & Compfs(1) & 0.35 & 1.0 & - & 100 & 30 & 0.003 & 0.99\\
            & LASSO & 0.4 & - & - & 50 & - & 0.003 & 0.99\\
            & Group LASSO & 0.04 & - & - & Full & - & - & - \\
            & STG & 0.1 & - & - & 250 & 20 & 0.001 & 1.00 \\
            & CAE & 3 & - & - & 250 & 200 & 0.001 & 1.00 \\
            \midrule
			\multirow{6}{*}{Syn4}
            & Compfs(5) & 4.5 & 1.0 & 1.2 & 50 & 20 & 0.003 & 0.99 \\
            & Compfs(1) & 0.35 & 1.0 & - & 100 & 30 & 0.003 & 0.99\\
            & LASSO & 0.4 & - & - & 50 & - & 0.003 & 0.99\\
            & Group LASSO & 0.04 & - & - & Full & - & - & - \\
            & STG & 0.1 & - & - & 250 & 20 & 0.001 & 1.00 \\
            & CAE & 4 & - & - & 250 & 200 & 0.001 & 1.00 \\
            \midrule
			\multirow{6}{*}{Chem1}
            & Compfs(5) & 2.0 & 1.0 & 1.2 & 20 & 20 & 0.003 & 0.99\\
            & Compfs(1) & 0.4 & 1.0 & - & 20 & 30 & 0.003 & 0.99\\
            & LASSO & 0.4 & - & - & 20 & - & 0.003 & 0.99\\
            & Group LASSO & 0.04 & - & - & Full & - & - & - \\
            & STG & 0.1 & - & - & 250 & 20 & 0.001 & 1.00 \\
            & CAE & 2 & - & - & 250 & 200 & 0.001 & 1.00 \\
            \midrule
            \multirow{6}{*}{Chem2}
            & Compfs(5) & 3.4 & 1.0 & 1.2 & 20 & 20 & 0.003 & 0.99\\
            & Compfs(1) & 0.4 & 1.0 & - & 20 & 30 & 0.003 & 0.99\\
            & LASSO &  0.2 & - & - & 20 & - & 0.003 & 0.99\\& Group LASSO & 0.04 & - & - & Full & - & - & - \\
            & STG & 0.1 & - & - & 250 & 20 & 0.001 & 1.00 \\
            & CAE & 3 & - & - & 250 & 200 & 0.001 & 1.00 \\
            \midrule
            \multirow{6}{*}{Chem3}
            & Compfs(5) & 2.0 & 1.0 & 1.2 & 20 & 20 & 0.003 & 0.99\\
            & Compfs(1) & 0.7 & 1.0 & - & 20 & 30 & 0.003 & 0.99\\
            & LASSO & 0.2 & - & - & 20 & - & 0.003 & 0.99\\
            & Group LASSO & 0.04 & - & - & Full & - & - & - \\
            & STG & 0.1 & - & - & 250 & 20 & 0.001 & 1.00 \\
            & CAE & 4 & - & - & 250 & 200 & 0.001 & 1.00 \\
			\bottomrule
        \end{tabular} 
 	\end{sc}
 	\end{small}
 	\end{center}
    \hfill
    \vskip -0.1in
\end{table*}

We train every model apart from Group LASSO with the ADAM optimizer \cite{kingma2014adam}, with the learning rates and decays given in Table \ref{Table:hyperparams}. \proposed(5) is trained for 35 epochs, \proposed(1) is trained for 35 epochs, LASSO is trained for 8 epochs and Group LASSO is trained for 1000 iterations or until convergence. The experiments were all repeated 10 times to produce means and standard deviations. For all experiments we use a threshold of 0.7 to determine whether a feature is relevant for \proposed. For LASSO, feature $i$ is relevant if $|w_i|>0.01$, where $\wv$ is the weight vector. If irrelevant, $w_i$ is set to 0 during evaluation. For Group LASSO, group $i$ is relevant if $||w_{\Gc_i}||_2>0.005$. If irrelevant, $w_{\Gc_i}$ is set to 0 during evaluation. When evaluating the composite features discovered by \proposed, we disregard repeated or empty groups, only considering unique non-empty ones.

\paragraph{Reproducibility.} 
Code is available at either of the following GitHub repositories: \url{https://github.com/a-norcliffe/Composite-Feature-Selection}, \url{https://github.com/vanderschaarlab/Composite-Feature-Selection}. Most experiments were run locally on a machine with 8GB RAM,  Intel Core i7-7500 HQ, NVIDIA GeForce MX150. The remainder (Group LASSO, STG, CAE, MNIST) were run on the free version of Google Colab.

\clearpage
\section{Jaccard Index}
\label{app:jaccard}

The Jaccard Index (or Jaccard Similarity) \cite{Jaccard1912} is a metric to estimate the similarity between two sets $A$ and $B$, which returns a scalar value between 0 and 1. It is defined as the ratio of the number of elements in the intersection and the number of elements in the union of the two sets:
\begin{equation}
\label{eq:jaccard_def}
    \mathcal{J}(A, B) = \frac{|A \cap B|}{|A \cup B|}
\end{equation}
When carrying out standard feature selection this would work as a metric to compare the set of ground truth features to the set of discovered features. However, in the new problem of composite feature selection both the ground truth and the output are a set of sets, i.e. a set of groups of features. If we were to use this metric on the sets $\{ \{1, 2\}, \{3, 4\}\}$ and $\{ \{ 1, 2, 3\}, \{4\}\}$, Eq. \eqref{eq:jaccard_def} would return 0, because the two sets of sets share no elements, where the elements are the composite features. This is clearly undesirable behaviour as we can see qualitatively the learnt group structure is similar to the true group structure. Therefore, we generalise the Jaccard Similarity to the Group Similarity denoted by G\textsubscript{Sim}:
\begin{equation}
\text{G\textsubscript{sim}} = \frac{1}{\max(N,K)} \sum_{i=1}^{N} \max_{j \in [K]} \mathcal{J}(\Gc_i, \hat{\Gc}_j) 
\end{equation}
Here we calculate the Jaccard Similarity between all the learnt composite features and the true composite features. For each true composite feature we look for the learnt composite feature that matches it best, and take the sum for all the true composite features. We then normalise this sum by dividing by either the number of true composites or the number of discovered composites, whichever is larger. This firstly keeps G\textsubscript{Sim} between 0 and 1, with $G\textsubscript{Sim}=0$ when none of the correct features are discovered, and 1 for perfect recovery of \emph{only} the true group structure. However, if the true group structure is incidentally discovered, (if many composites are discovered for example), then the normalisation term punishes any additional structure which is incorrect. This prevents a method finding all combinations of true individual features. Further to this, it will reward the group structures that are similar more than those that are significantly different. To help understand the group similarity metric further we provide some examples in Table \ref{Table:jaccard_example}.

\begin{table*}[h]
	\vskip -0.05in
	\caption{Examples of Group Similarity calculations.} \label{Table:jaccard_example}
	\vskip 0.15in
	\begin{center}
    \begin{small}
    \begin{sc}
		\begin{tabular}[b]{c c c c c}
			\toprule
			\textbf{True Composites} & \textbf{Candidate Composites} & \textbf{Sum} & \textbf{max}$\mathbf{(N, K)}$ &\textbf{G\textsubscript{Sim}} \\
			\midrule
			\multirow{5}{*}{ $\{1, 2\}, \{3, 4\}$ }
			& $\{1, 2\}, \{3, 4\}$ & 2 & 2 & 1
			\\
			& $\{1, 2, 3, 4\}$ & 1 & 2 & 1/2 \\
			& $\{1, 2, 3\}, \{1, 4\}$ & 1 & 2 & 1/2 \\
			& $\{1\}$, $\{2\}$, $\{3\}$, $\{4\}$ & 1 & 4 & 1/4\\
			& $\{1, 2\}, \{3, 4\}, \{1, 3\}, \{1, 4\}, \{2, 3\}, \{2, 4\}$ & 2 & 6 & 1/3\\
			\midrule
			\multirow{3}{*}{$\{1\}, \{2\}, \{3, 4, 5\}$}
			& $\{1, 2\}$ & 1 & 3 & 1/3
			\\
			& $\{3\}, \{1, 3, 5\}$ & 5/6 & 3 & 5/18\\
			& $\{6\}, \{7\}, \{8, 9, 10\}$ & 0 & 3 & 0\\
			\midrule
			\multirow{3}{*}{$\{1, 2\}, \{1, 3\}$} & $\{1, 2\}, \{1, 3\}$ & 2 & 2 & 1 \\
			& $\{1, 2, 3\}$ & 4/3 & 2 & 2/3\\
			& $\{1\}, \{2\}, \{3\}$ & 1 & 3 & 1/3 \\
			\bottomrule
        \end{tabular} 
 	\end{sc}
 	\end{small}
 	\end{center}
    \hfill
    \vskip -0.1in
\end{table*}

In order to provide a more complete picture of a model's group predictions, we also consider the feature level true positive and false discovery rates. This is analogous to using accuracy for classification as well as other metrics such as AUROC. 

\clearpage
\section{Chemistry Datasets}
\label{app:chem}

The semi-synthetic molecular datasets employed are based on several of the datasets constructed by \cite{McCloskey2019attribution}, which were released under the Apache 2.0 license.\footnote{Data from \url{https://github.com/google-research/graph-attribution/raw/main/data/all_16_logics_train_and_test.zip}.}
The synthetic “binding logics” are expressed as a combination of molecular fragments that must either be present or absent for binding to occur and this is used to label molecules from the ZINC database \cite{Irwin2012zinc}. The synthetic binding logics used in our paper are shown in Table \ref{Table:chem_logics} and cover a range of ground truth group structures.
The logics correspond to the following combinations of molecular fragments: 
(1) Ether OR NO Alkyne; 
(2) (Primary Amine or NO Benzene) AND NO Ether
(3) (Benzene AND NO Carbonyl) OR (Alkyne AND NO Ether).
These were chosen from the larger set of synethic binding logics considered by \cite{McCloskey2019attribution} since they exhibit a group structure.

In order to reduce possible biases, \cite{McCloskey2019attribution} constructed each logic dataset to be balanced across the set of all binary combinations. 
For example, given a logic $A \& B$, the datasets contain an equal number of examples for every logic combination: $A \& B$, $A \& \neg B$, $\neg A \& B$ and $\neg A \& \neg B$. 
However, this induces significant class imbalance for the more complex binding logics.
We used the same training and test sets determined by \cite{McCloskey2019attribution}. The number of examples varies for the different binding logics, with the training sets employed ranging from 3,862 to 14,769 molecules, and the test sets from 467 to 1832.

Molecules are featurized using a set of 84 functional groups, where feature $x_i=1$ if the molecule contains functional group $i$ and $0$ otherwise.
Ground truth indices in Table \ref{Table:chem_logics} correspond to the relevant indices in the molecular featurization.

\begin{table*}[ht]
	\vskip -0.05in
	\caption{Binding Logics for Chemistry Datasets.} \label{Table:chem_logics}
	\vskip 0.15in
	\centering
    \begin{small}
    \begin{sc}
		\begin{tabular}[b]{c c c}
			\toprule
			\textbf{Dataset} & \textbf{Binding Logic (Visualization)} &\textbf{ground truth} \\
			\midrule
			\multirow{3}{*}{Chem1} & \multirow{3}{*}{\includegraphics[height=0.75cm]{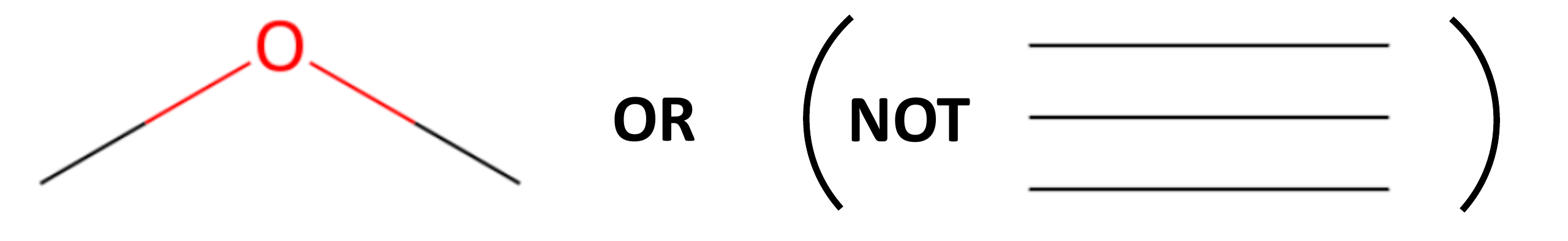}} & \multirow{3}{*}{\{40\}, \{1\}} \\
			\\
			\\
			\multirow{3}{*}{Chem2} & \multirow{3}{*}{\includegraphics[height=0.75cm]{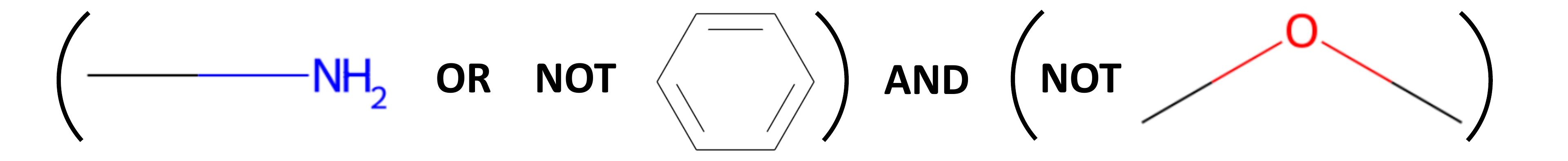}} & \multirow{3}{*}{\{56, 18\}, \{40\}} \\
			\\
			\\
			\multirow{3}{*}{Chem3} & \multirow{3}{*}{\includegraphics[height=0.75cm]{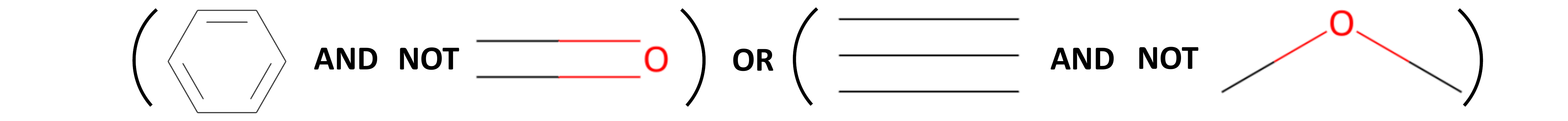}} & \multirow{3}{*}{\{18, 29\}, \{1, 40\}} \\
			\\
			\\
			\bottomrule
        \end{tabular} 
 	\end{sc}
 	\end{small}
    \hfill
    \vskip -0.1in
\end{table*}

\section{Additional Baselines}
\label{app:add_experiments}

Here we run further baselines on our experiments, including implementations details and hyperparameters as well as results. 

\paragraph{Reduced \proposed{}.} We include an ablation of our method, which uses only one group (\proposed(1)), which tests how well the method performs as a standard feature selection algorithm. Hyperparameters are given in Table \ref{Table:hyperparams}. Note that due to the loss function used to train our method (in particular Eq. \ref{eq:loss_individual}), we do not expected {\proposed(1)} to outperform state of the art tradition feature selection methods.

\paragraph{Unsupervised Concrete Autoencoder.} We consider the unsupervised version of concrete autoencoder (unsup-CAE), where the minimum subset of features are used to reconstruct the entire input. Unsup-CAE uses the same hyperparameters as Sup-CAE, we evaluate model performance by looking at the reconstruction MSE.

\paragraph{Random Forests.} Attempting to create grouping we use Random Forests. This model consists of a collection of 250 decision trees, with maximum depth 5. After training, the top 10 performing trees are considered. For each one the chosen features are those with a feature importance greater than 0.1. This collection of features is considered a group and all unique groups are used as the predicted composites. This was implemented using the scikitlearn implementation of Random Forest.

\paragraph{Gradient Boosted Decision Trees.} This is the same as the random forests benchmark, but here Gradient Boosted Decision Trees are used rather than simple Decision Trees. We refer to this method as GBDT. The same hyperparameters are used as Random Forests. This was implemented using the scikitlearn implementation of Gradient Boosted Decision Trees.

\paragraph{Clustering on Oracle Features.} For the synthetic and semi-synthetic tasks, we take the true features and apply clustering to them (Oracle+Cluster). This attempts to group them based on their correlations rather than interactions. In each case we enforce the same number of clusters as number of true composites. The clustering is done using scikitlearn Feature Agglomeration.

\paragraph{Ensemble of STGs.} Our final baseline is an ensemble of STGs (Ensemble STG). Here we use a collection of independent STGs. However, rather than their outputs being directly used as logits, they are instead all concatenated. Then this new larger hidden representation is passed through an MLP. The output of this is used to predict the logits which are used to train the ensemble. The regularisation loss remains the same. The features selected by each STG are then a composite feature. Our implementation uses two hidden layers for each individual STG and one hidden layer for the final MLP, ReLU activations are used. We use a hidden width of 50. For the tasks we train with 4 STGs, using $\lambda=0.1$ for synthetic tasks and $\lambda=4.0$ for the Chemistry tasks. We train using the ADAM optimizer with batchsizes of 100 for synthetic tasks and 500 for chemistry tasks, the learning rates are 0.001 for synthetic tasks and 0.01 for chemistry tasks.

Next we give the results of these new baselines alongside our main ones. Additional baselines are run 5 times.

\subsection{Synthetic Task}
\label{app:baselines}

We test all our benchmarks on the synthetic tasks from Section \ref{sec:exp}. The results are given in Table \ref{Table:Gauss_extended}, we present them along with the main results for easy comparison.

\proposed(1) performs well on the synthetic tasks, showing it is able to carry out feature selection.
STG performs as well as the Oracle on Syn1-3, demonstrating it as a state of the art feature selection method, however has high FDR on Syn4. Sup-CAE significantly underperforms other non-linear methods on Syn2 and Syn3, although performs well on Syn4, achieving the lowest FDR of all methods with non-zero TPR, due to having to select how many features are found. 
Unsup-CAE performs poorly across Syn1-3 because all features are independent, so it is impossible to reconstruct the full input from a subset. As expected, the reconstruction MSE almost exactly equals the variance of the standard normal distribution from which features are sampled. It performs slightly better on Syn4 due to correlation between features. 
Random Forests achieve perfect TPR, and on Syn 1-3 achieve reasonable group similarities compared to Oracle, showing it is possible to provide some grouping. However, due to the high correlation performs poorly on Syn 4. 
GBDT performs as well as Oracle on all tasks, However, whilst this is a strength through the lens of standard feature selection, it is a weakness in composite feature selection because no grouping has been provided. 
Oracle+Cluster trivially achieves the same TPR and FDR as Oracle. Additionally, it also occasionally achieves higher group similarity, this is because it does enforce some group structure. However, due to a lack of correlation between features, the grouping is inferior to that provided by \proposed{}. This is to be expected, as this groups based on correlations between features rather than their interactions. 
Ensemble STG performs the same as STG, showing that it has not been able to separate out the groups. 

Noticeably, \proposed{} still performs very well, performing either best or close to best in terms of discovering group structure as measured by G\textsubscript{Sim}, as well as on TPR and FDR.

\clearpage

\begin{table*}[ht]
	\vskip -0.05in
	\caption{Performance on Synthetic Datasets including all benchmarks, values are recorded with their standard deviations. Performance is recorded as accuracy except for Unsup-CAE, where the reconstruction MSE is given.} \label{Table:Gauss_extended}
	\vskip -0.05in
	\begin{center}
        \resizebox{\linewidth}{!}{
		\begin{tabular}[b]{c c c c c c c}
			\toprule
			\multirow{2}{*}{\textbf{Dataset}} & \multirow{2}{*}{\textbf{Model}} & \multirow{2}{*}{\textbf{TPR}} & \multirow{2}{*}{\textbf{FDR}} & \multirow{2}{*}{\textbf{G\textsubscript{sim}}} & \multirow{2}{*}{\textbf{No. Groups}} & \textbf{Accuracy} (\%)\\
			& & & & & &
			\textbf{Or MSE} \\
			\midrule
			\multirow{12}{*}{Syn1} 
			& CompFS(5) & 100.0 {\scriptsize $\pm$ 0.0} & 0.0 {\scriptsize $\pm$ 0.0} &  0.91 {\scriptsize $\pm$ 0.14} & 2.2 {\scriptsize $\pm$ 0.4} &  98.9 {\scriptsize $\pm$ 0.5}   \\
			 & Oracle & 100.0 {\scriptsize $\pm$ 0.0}  & 0.0 {\scriptsize $\pm$ 0.0}   &  0.50 {\scriptsize $\pm$ 0.00}  & 1.0 {\scriptsize $\pm$ 0.0} & 100.0 {\scriptsize $\pm$ 0.0} \\
			 & CompFS(1) & 100.0 {\scriptsize $\pm$ 0.0} & 0.0 {\scriptsize $\pm$ 0.0} &  0.50 {\scriptsize $\pm$ 0.00} & 1.0 {\scriptsize $\pm$ 0.0} &  99.1 {\scriptsize $\pm$ 0.8}   \\
			 & LASSO & 100.0 {\scriptsize $\pm$ 0.0}  & 0.0 {\scriptsize $\pm$ 0.0}   &  0.50 {\scriptsize $\pm$ 0.00}  & 1.0 {\scriptsize $\pm$ 0.0} & 81.8 {\scriptsize $\pm$ 2.0} \\
			 & Group LASSO & 100.0 {\scriptsize $\pm$ 0.0}  & 0.0 {\scriptsize $\pm$ 0.0}   &  0.67 {\scriptsize $\pm$ 0.00}  & 3.0 {\scriptsize $\pm$ 0.0} & 83.8 {\scriptsize $\pm$ 1.4} \\
			 & STG & 100.0 {\scriptsize $\pm$ 0.0} & 0.0 {\scriptsize $\pm$ 0.0} &  0.50 {\scriptsize $\pm$ 0.00} & 1.0 {\scriptsize $\pm$ 0.0} &  97.8 {\scriptsize $\pm$ 1.4}\\
			 & Sup-CAE & 100.0 {\scriptsize $\pm$ 0.0} & 0.0 {\scriptsize $\pm$ 0.0} &  0.50 {\scriptsize $\pm$ 0.00} & 1.0 {\scriptsize $\pm$ 0.0} &  97.8 {\scriptsize $\pm$ 1.4} \\
			 & Unsup-CAE & 0.0 {\scriptsize $\pm$ 0.0} & 100.0 {\scriptsize $\pm$ 0.0} &  0.00 {\scriptsize $\pm$ 0.00} & 1.0 {\scriptsize $\pm$ 0.0} &  0.999 {\scriptsize $\pm$ 0.005}\\
			 & {Random Forests} & {100.0 {\scriptsize $\pm$ 0.0}} & {0.0 {\scriptsize $\pm$ 0.0}} & {0.68 {\scriptsize $\pm$ 0.03}} & {2.8 {\scriptsize $\pm$ 0.4}} & {96.6 {\scriptsize $\pm$ 3.9}}\\
			 & {GBDT} & {100.0 {\scriptsize $\pm$ 0.0}} & {0.0 {\scriptsize $\pm$ 0.0}} & {0.50 {\scriptsize $\pm$ 0.00}} & {1.0 {\scriptsize $\pm$ 0.0}} & {100.0 {\scriptsize $\pm$ 0.0}} \\
			 & {Oracle+Cluster} & {100.0 {\scriptsize $\pm$ 0.0}} & {0.0 {\scriptsize $\pm$ 0.0}} & {0.50 {\scriptsize $\pm$ 0.00}} & {2.0 {\scriptsize $\pm$ 0.0}} & {100.0 {\scriptsize $\pm$ 0.0}}\\
			  & {Ensemble STG} & {100.0 {\scriptsize $\pm$ 0.0}} & {0.0 {\scriptsize $\pm$ 0.0}} & {0.50 {\scriptsize $\pm$ 0.00}} & {1.0 {\scriptsize $\pm$ 0.0}} & {93.5 {\scriptsize $\pm$ 0.8}}\\
			\midrule
			\multirow{12}{*}{Syn2} 
			& CompFS(5)  &  95.0 {\scriptsize $\pm$ 15.0} & 0.0 {\scriptsize $\pm$ 0.0}  &   0.90 {\scriptsize $\pm$ 0.20} & 1.8 {\scriptsize $\pm$ 0.4}	& 95.5 {\scriptsize $\pm$ 5.4} \\
			 & Oracle    & 100.0 {\scriptsize $\pm$ 0.0}  & 0.0 {\scriptsize $\pm$ 0.0}  &  0.50 {\scriptsize $\pm$ 0.00} & 1.0 {\scriptsize $\pm$ 0.0} & 100.0 {\scriptsize $\pm$ 0.0} \\
			 & CompFS(1)  &  85.0 {\scriptsize $\pm$ 22.9} & 2.0 {\scriptsize $\pm$ 6.0}  &   0.48 {\scriptsize $\pm$ 0.08} & 1.0 {\scriptsize $\pm$ 0.0}	& 90.2 {\scriptsize $\pm$ 8.1} \\
			 & LASSO & 0.0 {\scriptsize $\pm$ 0.0}  & 0.0 {\scriptsize $\pm$ 0.0}  &  0.00 {\scriptsize $\pm$ 0.00} & 0.0 {\scriptsize $\pm$ 0.0} & 52.6 {\scriptsize $\pm$ 2.9} \\
			 & Group LASSO & 0.0 {\scriptsize $\pm$ 0.0}  & 0.0 {\scriptsize $\pm$ 0.0}   &  0.00 {\scriptsize $\pm$ 0.00}  & 0.0 {\scriptsize $\pm$ 0.0} & 52.2 {\scriptsize $\pm$ 0.9} \\
			 & STG & 100.0 {\scriptsize $\pm$ 0.0} & 0.0 {\scriptsize $\pm$ 0.0} &  0.50 {\scriptsize $\pm$ 0.00} & 1.0 {\scriptsize $\pm$ 0.0} &  93.9 {\scriptsize $\pm$ 2.2}\\
			 & Sup-CAE & 37.5 {\scriptsize $\pm$ 31.7} & 42.5 {\scriptsize $\pm$ 44.2} &  0.24 {\scriptsize $\pm$ 0.20} & 1.0 {\scriptsize $\pm$ 0.0} &  61.9 {\scriptsize $\pm$ 12.8}\\
			 & Unsup-CAE & 0.0 {\scriptsize $\pm$ 0.0} & 100.0 {\scriptsize $\pm$ 0.0} &  0.00 {\scriptsize $\pm$ 0.00} & 1.0 {\scriptsize $\pm$ 0.0} &  0.995 {\scriptsize $\pm$ 0.005}\\
			 & {Random Forests} & {100.0 {\scriptsize $\pm$ 0.0}} & {0.0 {\scriptsize $\pm$ 0.0}} & {0.28 {\scriptsize $\pm$ 0.08}} & {6.8 {\scriptsize $\pm$ 1.5}} & {62.9 {\scriptsize $\pm$ 3.7}}\\
			 & {GBDT} & {100.0 {\scriptsize $\pm$ 0.0}} & {0.0 {\scriptsize $\pm$ 0.0}} & {0.54 {\scriptsize $\pm$ 0.08}} & {2.2 {\scriptsize $\pm$ 0.4}} & {99.0 {\scriptsize $\pm$ 0.0}} \\
			 & {Oracle+Cluster} & {100.0 {\scriptsize $\pm$ 0.0}} & {0.0 {\scriptsize $\pm$ 0.0}} & {0.58 {\scriptsize $\pm$ 0.00}} & {2.0 {\scriptsize $\pm$ 0.0}} & {100.0 {\scriptsize $\pm$ 0.0}}\\
			  & {Ensemble STG} & {100.0 {\scriptsize $\pm$ 0.0}} & {0.0 {\scriptsize $\pm$ 0.0}} & {0.50 {\scriptsize $\pm$ 0.00}} & {1.0 {\scriptsize $\pm$ 0.0}} & {86.9 {\scriptsize $\pm$ 0.7}}\\
			\midrule
			\multirow{12}{*}{Syn3} 
			& CompFS(5) & 100.0 {\scriptsize $\pm$ 0.0} & 0.0 {\scriptsize $\pm$ 0.0}  & 0.68 {\scriptsize $\pm$ 0.05} & 1.3 {\scriptsize $\pm$ 0.5} & 97.4 {\scriptsize $\pm$ 1.1} \\
			 & Oracle    & 100.0 {\scriptsize $\pm$ 0.0} & 0.0 {\scriptsize $\pm$ 0.0}  &  0.67 {\scriptsize $\pm$ 0.00}  & 1.0 {\scriptsize $\pm$ 0.0}    & 100.0 {\scriptsize $\pm$ 0.0}    \\
			 & CompFS(1)  &  96.7 {\scriptsize $\pm$ 10.0} & 2.5 {\scriptsize $\pm$ 7.5}  &   0.65 {\scriptsize $\pm$ 0.05} & 1.0 {\scriptsize $\pm$ 0.0}	& 94.5 {\scriptsize $\pm$ 3.7} \\
			 & LASSO & 0.0 {\scriptsize $\pm$ 0.0}  & 0.0 {\scriptsize $\pm$ 0.0}  &  0.00 {\scriptsize $\pm$ 0.00} & 0.0 {\scriptsize $\pm$ 0.0} & 56.5 {\scriptsize $\pm$ 4.0} \\
			 & Group LASSO & 0.0 {\scriptsize $\pm$ 0.0}  & 0.0 {\scriptsize $\pm$ 0.0}   &  0.00 {\scriptsize $\pm$ 0.00}  & 0.0 {\scriptsize $\pm$ 0.0} & 54.6 {\scriptsize $\pm$ 1.3} \\
			 & STG & 100.0 {\scriptsize $\pm$ 0.0} & 0.0 {\scriptsize $\pm$ 0.0} &  0.67 {\scriptsize $\pm$ 0.00} & 1.0 {\scriptsize $\pm$ 0.0} &  95.3 {\scriptsize $\pm$ 1.7}\\
			 & Sup-CAE & 23.3 {\scriptsize $\pm$ 31.6} & 66.7 {\scriptsize $\pm$ 47.1} &  0.23 {\scriptsize $\pm$ 0.31} & 1.0 {\scriptsize $\pm$ 0.0} &  62.6 {\scriptsize $\pm$ 12.6}\\
			 & Unsup-CAE & 0.0 {\scriptsize $\pm$ 0.0} & 100.0 {\scriptsize $\pm$ 0.0} &  0.00 {\scriptsize $\pm$ 0.00} & 1.0 {\scriptsize $\pm$ 0.0} &  0.999 {\scriptsize $\pm$ 0.005}\\
			 & {Random Forests} & {100.0 {\scriptsize $\pm$ 0.0}} & {0.0 {\scriptsize $\pm$ 0.0}} & {0.54 {\scriptsize $\pm$ 0.06}} & {3.6 {\scriptsize $\pm$ 0.5}} & {51.3 {\scriptsize $\pm$ 0.6}}\\
			 & {GBDT} & {100.0 {\scriptsize $\pm$ 0.0}} & {0.0 {\scriptsize $\pm$ 0.0}} & {0.67 {\scriptsize $\pm$ 0.00}} & {1.0 {\scriptsize $\pm$ 0.0}} & {98.5 {\scriptsize $\pm$ 0.0}} \\
			 & {Oracle+Cluster} & {100.0 {\scriptsize $\pm$ 0.0}} & {0.0 {\scriptsize $\pm$ 0.0}} & {0.50 {\scriptsize $\pm$ 0.00}} & {2.0 {\scriptsize $\pm$ 0.0}} & {100.0 {\scriptsize $\pm$ 0.0}}\\
			  & {Ensemble STG} & {100.0 {\scriptsize $\pm$ 0.0}} & {0.0 {\scriptsize $\pm$ 0.0}} & {0.67 {\scriptsize $\pm$ 0.00}} & {1.0 {\scriptsize $\pm$ 0.0}} & {84.6 {\scriptsize $\pm$ 1.5}}\\
			\midrule
			\multirow{12}{*}{Syn4} 
			& CompFS(5)   & 90.0 {\scriptsize $\pm$ 12.2} & 51.9 {\scriptsize $\pm$ 13.8} &  0.47 {\scriptsize $\pm$ 0.20} & 2.5 {\scriptsize $\pm$ 0.7} 			& 95.8 {\scriptsize $\pm$ 1.8} \\
			 & Oracle    & 100.0 {\scriptsize $\pm$ 0.0} & 0.0 {\scriptsize $\pm$ 0.0}  &  0.50 {\scriptsize $\pm$ 0.00} & 1.0 {\scriptsize $\pm$ 0.0} & 100.0 {\scriptsize $\pm$ 0.0} \\
			 & CompFS(1)  &  75.0 {\scriptsize $\pm$ 19.4} & 63.1 {\scriptsize $\pm$ 6.0}  &   0.18 {\scriptsize $\pm$ 0.03} & 1.0 {\scriptsize $\pm$ 0.0}	& 92.8 {\scriptsize $\pm$ 5.9} \\
			 & LASSO & 0.0 {\scriptsize $\pm$ 0.0}  & 0.0 {\scriptsize $\pm$ 0.0}  &  0.00 {\scriptsize $\pm$ 0.00} & 0.0 {\scriptsize $\pm$ 0.0} & 51.8 {\scriptsize $\pm$ 3.2} \\
			 & Group LASSO & 0.0 {\scriptsize $\pm$ 0.0}  & 10.0 {\scriptsize $\pm$ 31.6}   &  0.00 {\scriptsize $\pm$ 0.00}  & 0.1 {\scriptsize $\pm$ 0.3} & 53.0 {\scriptsize $\pm$ 1.1} \\
			 & STG & 100.0 {\scriptsize $\pm$ 0.0} & 66.7 {\scriptsize $\pm$ 0.0} &  0.17 {\scriptsize $\pm$ 0.00} & 1.0 {\scriptsize $\pm$ 0.0} &  94.2 {\scriptsize $\pm$ 2.1}\\
			 & Sup-CAE & 72.5 {\scriptsize $\pm$ 14.2} & 16.7 {\scriptsize $\pm$ 14.7} &  0.39 {\scriptsize $\pm$ 0.08} & 1.0 {\scriptsize $\pm$ 0.0} &  72.2 {\scriptsize $\pm$ 13.2}\\
			 & Unsup-CAE & 32.5 {\scriptsize $\pm$ 26.5} & 67.5 {\scriptsize $\pm$ 26.5} &  0.15 {\scriptsize $\pm$ 0.14} & 1.0 {\scriptsize $\pm$ 0.0} &  0.983 {\scriptsize $\pm$ 0.007} \\
			 & {Random Forests} & {100.0 {\scriptsize $\pm$ 0.0}} & {63.5 {\scriptsize $\pm$ 2.1}} & {0.12 {\scriptsize $\pm$ 0.04}} & {9.8 {\scriptsize $\pm$ 0.4}} & {82.1 {\scriptsize $\pm$ 0.7}}\\
			 & {GBDT} & {100.0 {\scriptsize $\pm$ 0.0}} & {0.0 {\scriptsize $\pm$ 0.0}} & {0.44 {\scriptsize $\pm$ 0.00}} & {3.0 {\scriptsize $\pm$ 0.0}} & {98.5 {\scriptsize $\pm$ 0.0}} \\
			 & {Oracle+Cluster} & {100.0 {\scriptsize $\pm$ 0.0}} & {0.0 {\scriptsize $\pm$ 0.0}} & {0.58 {\scriptsize $\pm$ 0.00}} & {2.0 {\scriptsize $\pm$ 0.0}} & {100.0 {\scriptsize $\pm$ 0.0}}\\
			  & {Ensemble STG} & {100.0 {\scriptsize $\pm$ 0.0}} & {66.7 {\scriptsize $\pm$ 0.0}} & {0.17 {\scriptsize $\pm$ 0.00}} & {1.0 {\scriptsize $\pm$ 0.0}} & {81.2 {\scriptsize $\pm$ 0.4}}\\
			\bottomrule
        \end{tabular} 
        }
 	\end{center}
    \hfill
    \vskip -0.1in
\end{table*}

\clearpage

\subsection{Chemistry Task}

Next we test all our benchmarks on the chemistry tasks from Section \ref{sec:exp}. The results are given in Table \ref{Table:CHEM_extended}, we present them along with the main results for easy comparison.

\begin{table*}[ht]
	\vskip -0.05in
	\caption{Performance on Chemistry Datasets, values are recorded with their standard deviations. Performance is recorded as accuracy except for Unsup-CAE, where the reconstruction MSE is given.} \label{Table:CHEM_extended}
	\vskip -0.05in
	\begin{center}
        \resizebox{\linewidth}{!}{
		\begin{tabular}[b]{c c c c c c c}
			\toprule
			\multirow{2}{*}{\textbf{Dataset}} & \multirow{2}{*}{\textbf{Model}} & \multirow{2}{*}{\textbf{TPR}} & \multirow{2}{*}{\textbf{FDR}} & \multirow{2}{*}{\textbf{G\textsubscript{sim}}} & \multirow{2}{*}{\textbf{No. Groups}} & \textbf{Accuracy} (\%)\\
			& & & & & &
			\textbf{Or MSE} \\
			\midrule
			\multirow{12}{*}{Chem1} 
			& CompFS(5)  &  100.0 {\scriptsize $\pm$ 0.0} & 0.0 {\scriptsize $\pm$ 0.0}  &  0.82 {\scriptsize $\pm$ 0.20} &  1.9 {\scriptsize $\pm$ 0.5} & 100.0 {\scriptsize $\pm$ 0.0} \\
			& Oracle    & 100.0 {\scriptsize $\pm$ 0.0}  & 0.0 {\scriptsize $\pm$ 0.0}  &  0.50 {\scriptsize $\pm$ 0.00} & 1.0 {\scriptsize $\pm$ 0.0} & 100.0 {\scriptsize $\pm$ 0.0} \\
			& CompFS(1) & 100.0 {\scriptsize $\pm$ 0.0} & 0.0 {\scriptsize $\pm$ 0.0} & 0.50 {\scriptsize $\pm$ 0.00} & 1.0 {\scriptsize $\pm$ 0.0} & 100.0 {\scriptsize $\pm$ 0.0}  \\
			& LASSO & 100.0 {\scriptsize $\pm$ 0.0} & 0.0 {\scriptsize $\pm$ 0.0} & 0.50 {\scriptsize $\pm$ 0.00} & 1.0 {\scriptsize $\pm$ 0.0} & 75.8 {\scriptsize $\pm$ 0.0} \\
			& Group LASSO & 100.0 {\scriptsize $\pm$ 0.0}  & 0.0 {\scriptsize $\pm$ 0.0}   &  0.67 {\scriptsize $\pm$ 0.00}  & 3.0 {\scriptsize $\pm$ 0.0} & 100.0 {\scriptsize $\pm$ 0.0} \\
			& STG & 100.0 {\scriptsize $\pm$ 0.0} & 0.0 {\scriptsize $\pm$ 0.0} &  0.50 {\scriptsize $\pm$ 0.00} & 1.0 {\scriptsize $\pm$ 0.0} &  100.0 {\scriptsize $\pm$ 0.0}\\
			& Sup-CAE & 50.0 {\scriptsize $\pm$ 0.0} & 0.0 {\scriptsize $\pm$ 0.0} &  0.50 {\scriptsize $\pm$ 0.00} & 1.0 {\scriptsize $\pm$ 0.0} &  75.8 {\scriptsize $\pm$ 0.0}\\
			 & Unsup-CAE & 0.0 {\scriptsize $\pm$ 0.0} & 100.0 {\scriptsize $\pm$ 0.0} &  0.00 {\scriptsize $\pm$ 0.00} & 1.0 {\scriptsize $\pm$ 0.0} &  0.042 {\scriptsize $\pm$ 0.003} \\
			 & {Random Forests} & {100.0 {\scriptsize $\pm$ 0.0}} & {20.0 {\scriptsize $\pm$ 16.3}} & {0.50 {\scriptsize $\pm$ 0.06}} & {1.6 {\scriptsize $\pm$ 0.5}} & {98.37 {\scriptsize $\pm$ 2.0}}\\
			 & {GBDT} & {100.0 {\scriptsize $\pm$ 0.0}} & {0.0 {\scriptsize $\pm$ 0.0}} & {0.50 {\scriptsize $\pm$ 0.00}} & {1.0 {\scriptsize $\pm$ 0.0}} & {100.0 {\scriptsize $\pm$ 0.0}} \\
			 & {Oracle+Cluster} & {100.0 {\scriptsize $\pm$ 0.0}} & {0.0 {\scriptsize $\pm$ 0.0}} & {1.00 {\scriptsize $\pm$ 0.00}} & {2.0 {\scriptsize $\pm$ 0.0}} & {100.0 {\scriptsize $\pm$ 0.0}}\\
			  & {Ensemble STG} & {100.0 {\scriptsize $\pm$ 0.0}} & {0.0 {\scriptsize $\pm$ 0.0}} & {0.50 {\scriptsize $\pm$ 0.00}} & {1.0 {\scriptsize $\pm$ 0.0}} & {100.0 {\scriptsize $\pm$ 0.0}}\\
			\midrule
			\multirow{12}{*}{Chem2} 
			& CompFS(5)   & 100.0 {\scriptsize $\pm$ 0.0}  &
			0.0 {\scriptsize $\pm$ 0.0} &  0.72 {\scriptsize $\pm$ 0.24} & 2.2 {\scriptsize $\pm$ 0.6}
			& 100.0 {\scriptsize $\pm$ 0.0} \\
			& Oracle    & 100.0 {\scriptsize $\pm$ 0.0} & 0.0 {\scriptsize $\pm$ 0.0}  &  0.50 {\scriptsize $\pm$ 0.00} & 1.0 {\scriptsize $\pm$ 0.0} & 100.0 {\scriptsize $\pm$ 0.0} \\
			& CompFS(1) & 100.0 {\scriptsize $\pm$ 0.0} & 0.0 {\scriptsize $\pm$ 0.0}  &  0.50 {\scriptsize $\pm$ 0.00} & 1.0 {\scriptsize $\pm$ 0.0} & 100.0 {\scriptsize $\pm$ 0.0} \\
			& LASSO & 100.0 {\scriptsize $\pm$ 0.0} & 0.0 {\scriptsize $\pm$ 0.0} & 0.50 {\scriptsize $\pm$ 0.00} & 1.0 {\scriptsize $\pm$ 0.0} & 81.6 {\scriptsize $\pm$ 0.0} \\
			& Group LASSO & 100.0 {\scriptsize $\pm$ 0.0}  & 0.0 {\scriptsize $\pm$ 0.0}   &  0.40 {\scriptsize $\pm$ 0.00}  & 5.0 {\scriptsize $\pm$ 0.0} & 81.6 {\scriptsize $\pm$ 0.0} \\
			& STG & 100.0 {\scriptsize $\pm$ 0.0} & 0.0 {\scriptsize $\pm$ 0.0} &  0.50 {\scriptsize $\pm$ 0.00} & 1.0 {\scriptsize $\pm$ 0.0} &  100.0 {\scriptsize $\pm$ 0.0}\\
			& Sup-CAE & 66.7 {\scriptsize $\pm$ 0.0} & 0.0 {\scriptsize $\pm$ 0.0} &  0.42 {\scriptsize $\pm$ 0.00} & 1.0 {\scriptsize $\pm$ 0.0} &  80.9 {\scriptsize $\pm$ 9.5}\\
			 & Unsup-CAE & 0.0 {\scriptsize $\pm$ 0.0} & 100.0 {\scriptsize $\pm$ 0.0} &  0.00 {\scriptsize $\pm$ 0.00} & 1.0 {\scriptsize $\pm$ 0.0} &  0.040 {\scriptsize $\pm$ 0.003} \\
			 & {Random Forests} & {100.0 {\scriptsize $\pm$ 0.0}} & {25.0 {\scriptsize $\pm$ 0.0}} & {0.47 {\scriptsize $\pm$ 0.04}} & {2.4 {\scriptsize $\pm$ 0.5}} & {85.1 {\scriptsize $\pm$ 0.5}}\\
			 & {GBDT} & {100.0 {\scriptsize $\pm$ 0.0}} & {0.0 {\scriptsize $\pm$ 0.0}} & {0.50 {\scriptsize $\pm$ 0.00}} & {1.0 {\scriptsize $\pm$ 0.0}} & {100.0 {\scriptsize $\pm$ 0.0}} \\
			 & {Oracle+Cluster} & {100.0 {\scriptsize $\pm$ 0.0}} & {0.0 {\scriptsize $\pm$ 0.0}} & {0.50 {\scriptsize $\pm$ 0.00}} & {2.0 {\scriptsize $\pm$ 0.0}} & {100.0 {\scriptsize $\pm$ 0.0}} \\
			  & {Ensemble STG} & {100.0 {\scriptsize $\pm$ 0.0}} & {0.0 {\scriptsize $\pm$ 0.0}} & {0.50 {\scriptsize $\pm$ 0.00}} & {1.0 {\scriptsize $\pm$ 0.0}} & {100.0 {\scriptsize $\pm$ 0.0}}\\
			 \midrule
			\multirow{12}{*}{Chem3} 
			& CompFS(5)   & 100.0 {\scriptsize $\pm$ 0.0} & 7.3 {\scriptsize $\pm$ 11.7} &  0.62 {\scriptsize $\pm$ 0.17} & 2.4  {\scriptsize $\pm$ 0.5}
			& 100.0 {\scriptsize $\pm$ 0.0} \\
			& Oracle    & 100.0 {\scriptsize $\pm$ 0.0} & 0.0 {\scriptsize $\pm$ 0.0}  &  0.50 {\scriptsize $\pm$ 0.00} & 1.0 {\scriptsize $\pm$ 0.0} & 100.0 {\scriptsize $\pm$ 0.0} \\
			& CompFS(1) & 100.0 {\scriptsize $\pm$ 0.0} & 3.3 {\scriptsize $\pm$ 10.0} & 0.48 {\scriptsize $\pm$ 0.05} & 1.0 {\scriptsize $\pm$ 0.0} & 100.0 {\scriptsize $\pm$ 0.0} \\
			& LASSO & 100.0 {\scriptsize $\pm$ 0.0} & 0.0 {\scriptsize $\pm$ 0.0} & 0.50 {\scriptsize $\pm$ 0.00} & 1.0 {\scriptsize $\pm$ 0.0} & 87.4 {\scriptsize $\pm$ 5.2} \\
			& Group LASSO & 100.0 {\scriptsize $\pm$ 0.0}  & 20.0 {\scriptsize $\pm$ 0.0}   &  0.20 {\scriptsize $\pm$ 0.00}  & 10.0 {\scriptsize $\pm$ 0.0} & 91.5 {\scriptsize $\pm$ 0.0} \\
			& STG & 100.0 {\scriptsize $\pm$ 0.0} & 0.0 {\scriptsize $\pm$ 0.0} &  0.50 {\scriptsize $\pm$ 0.00} & 1.0 {\scriptsize $\pm$ 0.0} &  100.0 {\scriptsize $\pm$ 0.0}\\
			& Sup-CAE & 62.5 {\scriptsize $\pm$ 13.2} & 23.3 {\scriptsize $\pm$ 17.5} &  0.37 {\scriptsize $\pm$ 0.07} & 1.0 {\scriptsize $\pm$ 0.0} &  77.8 {\scriptsize $\pm$ 11.0}\\
			 & Unsup-CAE & 25.0 {\scriptsize $\pm$ 16.7} & 72.5 {\scriptsize $\pm$ 17.1} &  0.11 {\scriptsize $\pm$ 0.07} & 1.0 {\scriptsize $\pm$ 0.0} &  0.029 {\scriptsize $\pm$ 0.002} \\
			 & {Random Forests} & {100.0 {\scriptsize $\pm$ 0.0}} & {46.8 {\scriptsize $\pm$ 5.2}} & {0.15 {\scriptsize $\pm$ 0.01}} & {8.0 {\scriptsize $\pm$ 0.9}} & {95.9 {\scriptsize $\pm$ 1.7}}\\
			 & {GBDT} & {100.0 {\scriptsize $\pm$ 0.0}} & {0.0 {\scriptsize $\pm$ 0.0}} & {0.50 {\scriptsize $\pm$ 0.00}} & {1.0 {\scriptsize $\pm$ 0.0}} & {100.0 {\scriptsize $\pm$ 0.0}} \\
			 & {Oracle+Cluster} & {100.0 {\scriptsize $\pm$ 0.0}} & {0.0 {\scriptsize $\pm$ 0.0}} & {0.58 {\scriptsize $\pm$ 0.00}} & {2.0 {\scriptsize $\pm$ 0.0}} & {100.0 {\scriptsize $\pm$ 0.0}} \\
			  & {Ensemble STG} & {100.0 {\scriptsize $\pm$ 0.0}} & {0.0 {\scriptsize $\pm$ 0.0}} & {0.58 {\scriptsize $\pm$ 0.00}} & {2.0 {\scriptsize $\pm$ 0.0}} & {100.0 {\scriptsize $\pm$ 0.0}}\\
			\bottomrule
        \end{tabular} 
        }
 	\end{center}
    \hfill
    \vskip -0.2in
\end{table*}

STG performs as well as Oracle, further demonstrating its ability as a standard feature selection method. However, like Oracle, it does not offer any further insight into the structure of the discovered features. Sup-CAE does not perform well as it often fails to find all ground truth features. As in the synthetic tasks, Unsup-CAE performs poorly because the ability to reconstruct (we see from the MSE that this is achieved) does not equate to being able to predict, and therefore it finds non-predictive features. Due to high correlation between the functional gorups, Random Forests has the highest FDR (apart from Unsup-CAE), but it also achieves perfect TPR. As a result of many incorrect features being found, Random Forests has low group similarity. GBDT does as well as Oracle, however, again it only finds a single group, providing no information on the grouping. Orace+Cluster is able to achieve a group similarity higher than Oracle on Chem3 due to adding grouping, however it is not as high as \proposed{} due to grouping according to correlations rather than interactions. Ensemble STG performs in the same way as on the synthetic tasks. It achieves the same results as STG, but is unable to provide any grouping, except on Chem3, where some grouping was discovered, In this case it grouped all 4 features in one group and 3 of these in a second group, which is not the correct group structure, hence has lower G\textsubscript{sim} than \proposed.

Just as in the synthetic tasks, we see that \proposed{} consistently achieves the highest G\textsubscript{sim}, showing it is able to separate features into groups based on their interactions.

These additional benchmarks show us that even when it is possible to construct a model that can return groups of features, the groups are often not correct. However, \proposed{} is constantly able to achieve high G\textsubscript{sim}, showing the particular loss function and architecture work well for discovering composite features.

\subsection{MNIST}
\label{app:images}

We investigate \proposed{} on the MNIST dataset \cite{lecun1998gradient}, where the ground truth group structure is \emph{unknown}. Typically in images, pixels do not have any specific meaning; this is because we can translate an image slightly and the image is effectively unchanged to a human, but the pixels have entirely different values. We use MNIST as a qualitative example because the images are all centered and at the same scale. We consider \proposed{}, STG, Supervised CAE and Unsupervised CAE, all discovered pixels are given in Figure \ref{fig:mnist_discovered}. 

\paragraph{Implementation.} To train \proposed{}, we use 4 learners, with a hidden width of 256, temperature of 0.1, $\beta_R=\beta=0.18$. To select the features in each groups, we choose the top 15 features, those with the 15 highest $\pi$ values. The model was trained for 50 epochs with batchsize 500 and learning rate 0.001. The learnt groups each contain 15 features respectively with only 2 shared between any of the four, making 58 in total. For STG, we use $\lambda=4.0$ a batchsize of 500 and train with the ADAM optimizer with learning rate 0.001. STG discovered 46 pixels. For CAE we limit it to find 45 pixels, we train with a batchsize of 250 using the ADAM optimizer with learning rate 0.001.

We see what we would expect, all feature selection models find pixels near the centre and spread out, to sample as much of the image as possible.

\paragraph{Evaluation.} To fairly evaluate the discoveries, we train new models that use the discovered pixels as input only. For the individual groups, their union and the baselines we use an MLP with ReLU activations, two hidden layers and a hidden width of 200. For CompFS, we use the CompFS architecture as described in Appendix \ref{app:architecture}. We use a hidden width of 50 to make the number of parameters comparable to the other baselines. We train using the ADAM optimizer with a learning rate of 0.001 and batchsize of 250 for 30 epochs. The final accuracies are given in Table \ref{Table:mnist_accuracies}.

\begin{table*}[h]
	\vskip -0.05in
	\caption{Accuracy on the MNIST dataset.} \label{Table:mnist_accuracies}
	\vskip 0.15in
	\begin{center}
    \begin{small}
    \begin{sc}
		\begin{tabular}[b]{c c c}
			\toprule
			\textbf{Model} & \textbf{No. Pixels} & \textbf{Accuracy} \\
			\midrule
			Group 1 & 15 & 78.00\% \\
			Group 2 & 15 & 74.00\% \\
			Group 3 & 15 & 72.00\% \\
			Group 4 & 15 & 81.00\% \\
			CompFS & 58 & 94.00\% \\
			Union & 58 & 95.00\%  \\
			STG & 46 & 95.00\% \\
			Sup-CAE & 45 & 95.00\% \\
			Unsup-CAE & 45 & 95.00\% \\
			\bottomrule
        \end{tabular} 
 	\end{sc}
 	\end{small}
 	\end{center}
    \hfill
    \vskip -0.1in
\end{table*}

We see that the individual groups have low accuracies with the highest being group 4 at 81\%. Models trained on the union of these features, as well as those found by STG and CAE all achieve an accuracy of 95\%. Noteably, CompFS performs almost as well as the other baselines, despite having fewer parameters and a significantly constrained architecture. This demonstrates that the grouping does not drastically affect performance, despite enforcing constraints on how the features are used by the model.

\begin{figure}[h]
    \centering
    \includegraphics[width=\textwidth]{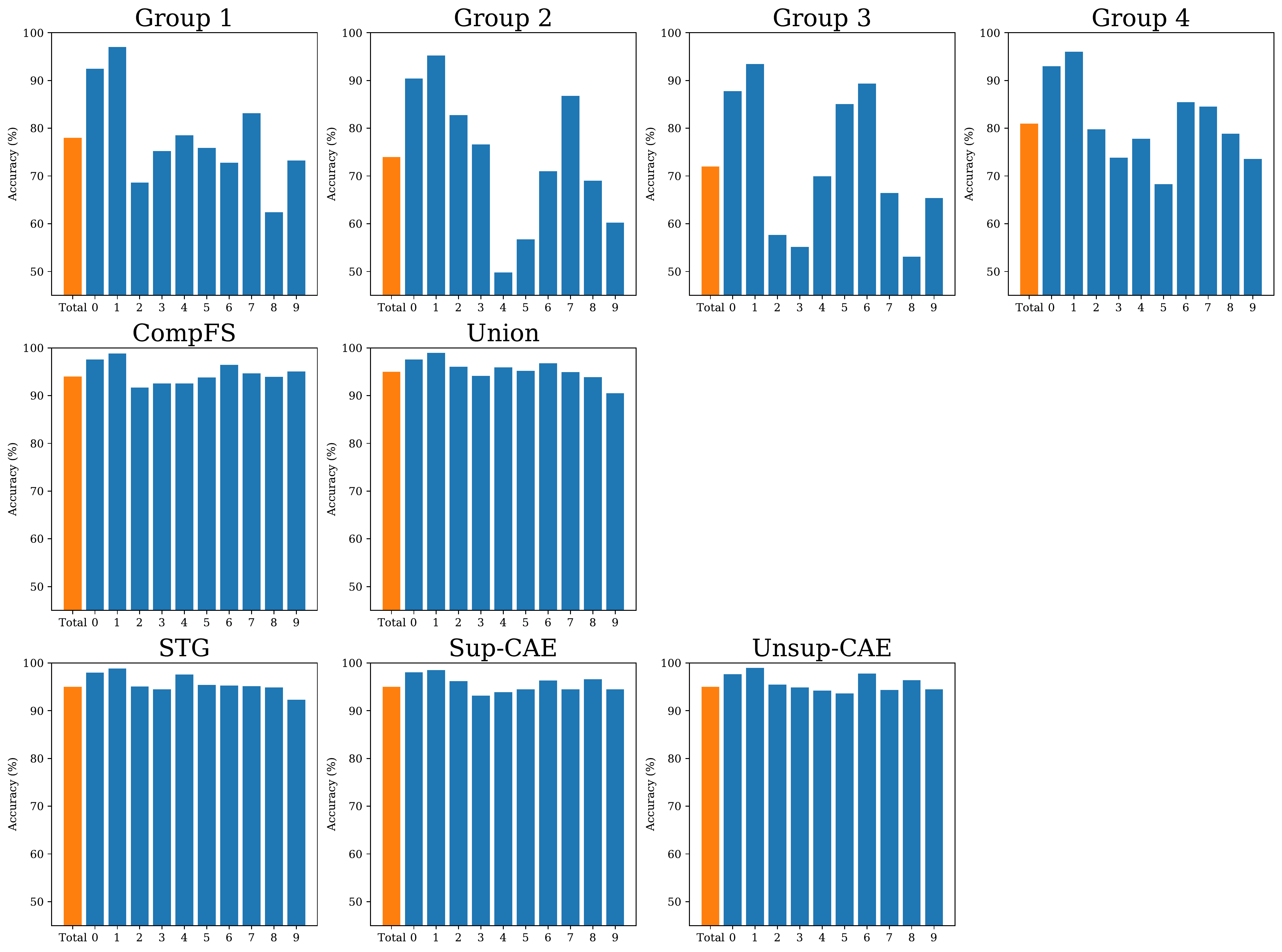}
    \caption{Class-level and overall MNIST accuracies of classifiers trained with the features found by each model. Note the scale of the y-axis does not start at 0, but at 45.}
    \label{fig:mnist_bar}
\end{figure}

Finally, we consider the class level accuracies of each of these models. We plot the total test accuracies and the class-level test accuracies in Figure \ref{fig:mnist_bar}. We see that in all groups there are digits which are not accurately classified. Whereas \proposed, STG and CAE are all able to recognise digits relatively uniformly. This shows that despite individual groups performing poorly, each one can be seen as providing some new information, and the ensemble combines this to make an overall prediction.

\clearpage

\section{METABRIC} \label{app:metabric}

In this section we provide additional details validating the features identified by {\proposed} on the METABRIC dataset.\footnote{Data from \url{https://www.kaggle.com/datasets/raghadalharbi/breast-cancer-gene-expression-profiles-metabric}}

\paragraph{Individual features.}
We first provide validation for the individual features discovered by {\proposed} in Table \ref{Table:METABRIC_features}. We found supporting evidence for 24 of the 25 features discovered by {\proposed}.

\begin{table*}[h]
	\vskip -0.05in
	\caption{Groups discovered by {\proposed} on the METABRIC dataset.} \label{Table:METABRIC_features}
	\vskip 0.15in
	\begin{center}
    \begin{small}
    \begin{sc}
		\begin{tabular}[b]{c c c c c c c c c}
			\toprule
			\textbf{Group} & \textbf{Feature} & \textbf{Ref.} &
			\textbf{Group} & \textbf{Feature} & \textbf{Ref.} & 
			\textbf{Group} & \textbf{Feature} & \textbf{Ref.} \\
			\midrule
			\multirow{5}{*}{Group 1} & psenen & \cite{Peltonen2013} &
			\multirow{5}{*}{Group 2} & bmp6   & \cite{Takahashi2008} &
			\multirow{5}{*}{Group 3} & bmpr2  & \cite{Pickup2015}
			\\
			& cxcr1 & \cite{Ginestier2010} &
			& mapk1 & \cite{Lu2019} & 
			& mmp12 & \cite{Cheng2022}
			\\
			& dlec1 & \cite{ALSARAKBI2010} &
			& smad3 & \cite{singha2019increased} &
			& asxl2 & \cite{Park2016}
			\\
			& mmp15 & \cite{Cheng2022} &
			& mapt  & \cite{Wang2019} &
			& birc6 & \cite{GomezBergna2021}
			\\
			& srd5a3 & \cite{Zhang2021} &
			& prkcq  & \cite{Byerly2020} &
			& star & \cite{Manna2019}
			\\
			\midrule
			\multirow{5}{*}{Group 4} & cdkn1a & \cite{wei2015expression} &
			\multirow{5}{*}{Group 5} & bmp4   & \cite{AMPUJA2016} &
			& & 
			\\
			& fgfr1 & \cite{Turner2010} &
			& mmp10 & \cite{Piskor2020} &
			& & 
			\\
			& tgfbr3 & \cite{Mei2007} &
			& tbl1xr1 & \cite{Li2014} &
			& & 
			\\
			& npnt & \cite{Wang2018} &
			& ush2a & &
			& & 
			\\
			& akr1c3 & \cite{ZHONG2015} &
			& hsd17b1 & \cite{Paivi2018} & 
			& & 
			\\
			\bottomrule
        \end{tabular} 
 	\end{sc}
 	\end{small}
 	\end{center}
    \hfill
    \vskip -0.1in
\end{table*}

Crucially, within each group, we found evidence of the joint importance of features, validating the groups discovered by CompFS.
Discovered relationships are provided in Table \ref{Table:METABRIC_pairs}.

\begin{table*}[h]
	\vskip -0.05in
	\caption{Supporting evidence for interactions between features within groups discovered by {\proposed} on the METABRIC dataset.} \label{Table:METABRIC_pairs}
	\vskip 0.15in
	\begin{center}
    \begin{small}
    \begin{sc}
		\begin{tabular}[b]{c c c}
			\toprule
			\textbf{Group} & \textbf{Features} & \textbf{Ref.}
			\\
			\midrule
			Group 1
			& psenen, cxcr1 & \cite{Bakele2014} \\
			\midrule
			\multirow{5}{*}{Group 2}
			& bmp6, smad3   & \cite{Chen2022} \\
			& bmp6, mapk1   & \cite{Zhang2018} \\
			& mapk1, smad3  & \cite{Fang2012} \\
			& mapk1, mapt   & \cite{Leugers2013} \\
			& mapk1, prkcq  & \cite{Byerly2016} \\
			\midrule
			Group 3
			& asxl2, birc6  & \cite{Kim2015} 
			\\
			\midrule
			\multirow{2}{*}{Group 4}
			& fgfr1, tgfbr3 & \cite{Wendt2014} \\
			& fgfr1, npnt   & \cite{Kato2018} \\
			\midrule
		    \multirow{2}{*}{Group 5}
			& bmp4, mmp10 & \cite{Fessing2010} \\
			& bmp4, hsd17b1 & \cite{Arcuri2021} \\
			\bottomrule
        \end{tabular} 
 	\end{sc}
 	\end{small}
 	\end{center}
    \hfill
    \vskip -0.1in
\end{table*}

\end{document}